\documentclass{article} % For LaTeX2e
\usepackage{iclr2025_conference,times}

\usepackage{amsmath,amsfonts,bm}

\def\eqref#1{equation~\ref{#1}}
\def\1{\bm{1}}

\DeclareMathAlphabet{\mathsfit}{\encodingdefault}{\sfdefault}{m}{sl}
\SetMathAlphabet{\mathsfit}{bold}{\encodingdefault}{\sfdefault}{bx}{n}

\usepackage{hyperref}
\usepackage{amsmath}
\usepackage{amssymb}
\usepackage{url}
\usepackage{graphicx}
\usepackage{algorithm}
\usepackage{algorithmic}
\usepackage{float}
\usepackage{multirow}
\usepackage{wrapfig}

\newtheorem{theorem}{Theorem}

\newtheorem{notations}{Notations}
\newtheorem{proposition}{Proposition}
\newtheorem{assumption}{Assumption}

\usepackage{tabularray}
\UseTblrLibrary{diagbox}

\usepackage[normalem]{ulem}

\title{Less is More: Masking Elements in Image Condition Features Avoids Content Leakages in Style Transfer Diffusion Models}

\author{Lin Zhu\textsuperscript{\rm1}, Xinbing Wang\textsuperscript{\rm1}, Chenghu Zhou\textsuperscript{\rm2}, Qinying Gu\textsuperscript{\rm3}, Nanyang Ye\textsuperscript{\rm1} \thanks{Nanyang Ye is the corresponding author. The code is available at \url{https://github.com/LinLLLL/MaskST}.}\\
\textsuperscript{\rm1} Shanghai Jiao Tong University, \textsuperscript{\rm2} Chinese Academy of Sciences,\\ 
\textsuperscript{\rm3} Shanghai Artificial Intelligence Laboratory\\
\texttt{\{zhulin\_sjtu, xwang8, ynylincoln\}@sjtu.edu.cn}, \\
\texttt{zhouchsjtu@gmail.com}, 
\texttt{guqinying@pjlab.org.cn} \\
}

\iclrfinalcopy % Uncomment for camera-ready version, but NOT for submission.
\begin{document}

\maketitle

\begin{abstract}
Given a style-reference image as the additional image condition, text-to-image diffusion models have demonstrated impressive capabilities in generating images that possess the content of text prompts while adopting the visual style of the reference image. However, current state-of-the-art methods often struggle to disentangle content and style from style-reference images, leading to issues such as \textit{content leakages}. To address this issue, we propose a masking-based method that efficiently decouples content from style without the need of tuning any model parameters. 
By simply masking specific elements in the style reference's image features,
we uncover a critical yet under-explored principle: guiding with appropriately-selected fewer conditions (e.g., dropping several image feature elements) can efficiently avoid unwanted content flowing into the diffusion models, enhancing the style transfer performances of text-to-image diffusion models.
In this paper, we validate this finding both theoretically and experimentally. Extensive experiments across various styles demonstrate the effectiveness of our masking-based method and support our theoretical results.
\end{abstract}

\section{Introduction}

\begin{figure}[H]
  \centering
  \includegraphics[width=1.0\linewidth]{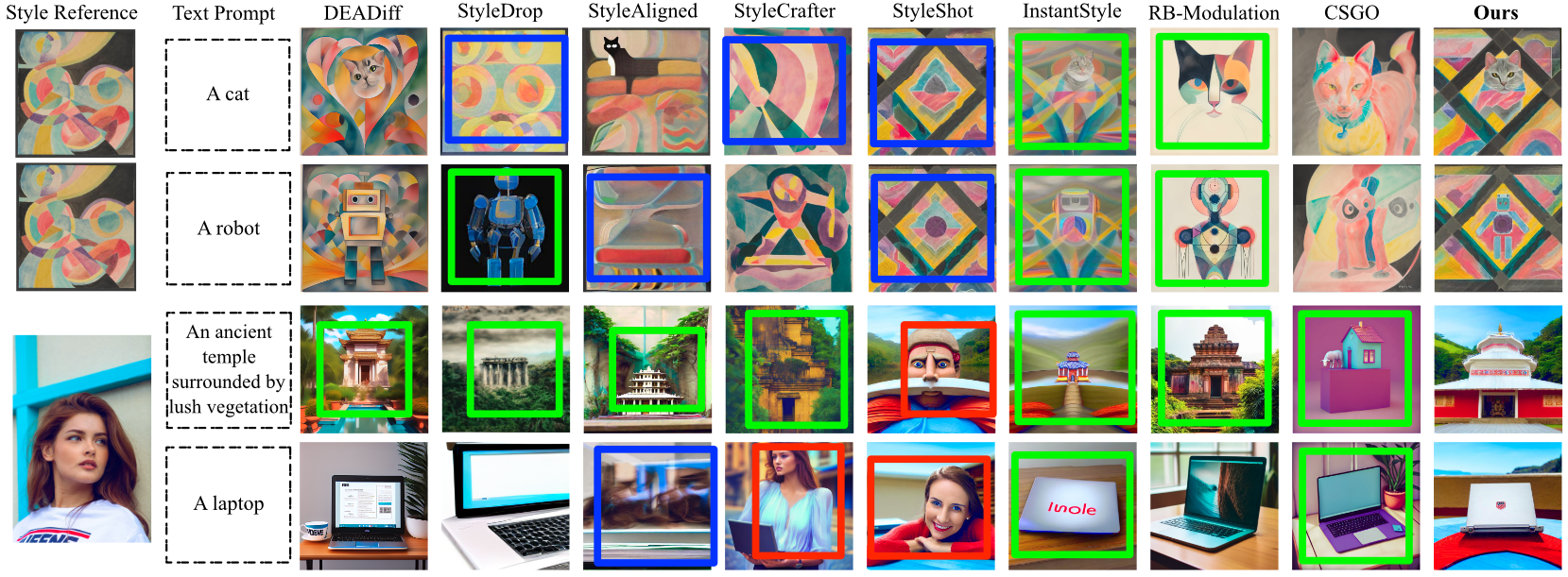}
  \vspace{-0.6cm}
   \caption{
   Given a style-reference image, our method is capable of synthesizing new images that resemble the style and are faithful to text prompts simultaneously. 
   Previous methods often face issues of either content leakages or style degradation.
   We mark the results with significant \textcolor{red}{content leakages}, \textcolor{green}{style degradation}, and \textcolor{blue}{loss of text fidelity} with \textcolor{red}{red}, \textcolor{green}{green}, and \textcolor{blue}{blue} boxes, respectively.
   }
   % \vspace{-0.3cm}
\label{fig:1}
\end{figure}

Recently, text-to-image diffusion models \citep{zhao2024uni, saharia2022photorealistic, zhang2023text, zhu2023conditional, zhang2023adding} have achieved notable success in generating high-quality images, especially for tasks requiring personalized image creation that preserves specific stylistic elements. By incorporating a style-reference image as an additional input, recent approaches \citep{park2024text, hamazaspyan2023diffusion, chung2024style, StyleAdapter, qi2024deadiff, InST} have effectively synthesized images that not only align with the content described in text prompts but also adopt the visual style of the reference image.
However, despite these advancements, content leakages from style-reference images remain a persistent issue \citep{wang2024instantstyle, ruiz2023dreambooth, stylepromptingswapping}, as illustrated in Figure~\ref{fig:1}.
\textit{Content leakages} occurs when intensifying the style transfer causes unintended non-stylistic elements from the reference image to be incorporated into the generated output \citep{wang2024instantstyle}. 
Conversely, reducing style intensity to prevent content leakage can result in \textit{style degradation}, hindering effective style transfer \citep{stylepromptingswapping}. 
{These challenges highlight the difficulty of disentangling styles from contents in style-reference images.}

Some approaches \citep{zhang2018separating, zhang2018unifiedframeworkgeneralizablestyle, qi2024deadiff} try to achieve the disentanglement by constructing paired datasets in which images share the same subject but exhibit distinct styles, facilitating the extraction of disentangled style and content representations. 
Other works \citep{sohn2024styledrop, liu2023stylecrafter, zhang2024artbank} optimize some or all of the model parameters using large sets of diverse style images, allowing them to isolate and integrate these stylistic elements into diffusion models. However, due to the inherently ambiguous nature of style, building comprehensive style datasets is resource-intensive and limits the model’s capacity to generalize to styles not present in the dataset. To address this issue, InstantStyle \citep{wang2024instantstyle} proposed a training-free strategy to separate style from content by subtracting content-related features from image features. Although this approach is simple and training-free, 
feature subtraction across different modalities inevitably introduces the image-text misalignment issue \citep{Misalign, gordon2023mismatch}, which hinders accurate disentanglement of content and style. As illustrated in Figure~\ref{fig:1}, although InstantStyle mitigates content leakage, it comes at the cost of significant style degradation.

% In the previous study, InstantStyle \citep{wang2024instantstyle}, it has been pointed out that tuning guidance strength weights between the image reference and text prompt can solve the content leakages problem. However, coefficient tuning is a very tricky thing and does not always work.
% To mitigate the content leakages issue, InstantStyle further proposed a feature subtraction method, which performed the subtraction operation of image features and content text features.
% However, we found out that the effectiveness of the feature subtraction method relies on coefficient tuning style strength. Users need to perform a labour-intensive coefficient-tuning process to achieve a balanced synthesis between content and style attributes. 
% According to its experiment results, at the beginning without subtraction, although the style strength has been manually adjusted, there are still certain content leaks. However, as the intensity of subtraction increases, the consistency between the generated results and the text is significantly enhanced, and there is no longer any content.

To overcome all these limitations, we propose a simple and effective training-free method that efficiently decouples content from style, without requiring tuning any model parameters.
Unlike InstantStyle, which subtracts features across different modalities, 
our approach removes content from the style-reference image by masking the image feature elements associated with the content. Specifically, we identify these content-related elements through clustering the element-wise product of the style-reference image features and the content text features, and then set their values to zero.
% The specific elements are identified based on their relevance to the content of reference images, with relevance determined by clustering the element-wise product of the style reference image features and the content text features. 
The theoretical evidence for the effectiveness of this identification approach is presented in Proposition~\ref{proposition}.

By simply masking specific elements in the style reference’s image features, we uncover a critical yet under-explored principle: guiding with appropriate masked conditions (e.g., masking several image feature elements) can prevent undesired content information from leaking into diffusion models, thereby improving style transfer performance.
We further present theoretical evidence for this principle.
As demonstrated in Theorem~\ref{theorem:1} and Theorem~\ref{theorem:2}, diffusion models guided by fewer appropriately selected conditions (e.g., the masked image feature and the text feature) achieve a lower divergence between the generated and real image distributions compared to models relying on more conditions that are less coherent (e.g., unfiltered image features combined with text and additional content features). This result aligns with the concept that ``Less is more''.
Extensive experiments across various styles, along with comparisons to state-of-the-art methods, validate the effectiveness of our approach and support our theoretical findings.

\section{Preliminaries}
\label{sec:Preliminaries}
Because of the portability and efficiency, we illustrate the proposed masking-based method based on the baseline module IP-Adapter \citep{ye2023ip}. 
In this section, we present the background knowledge and key observations from our initial experiments as follows:

\textbf{Conditional Diffusion Models} Diffusion models consist of two processes: a diffusion process (forward process), which incrementally adds Gaussian noise $\boldsymbol{\epsilon}$ to the data $\boldsymbol{x}_0$ through a Markov chain. Additionally, a denoising process generates samples from Gaussian noise $\boldsymbol{x}_T\sim N(0,1)$ with a learnable denoising model $\hat{\boldsymbol{\epsilon}}_\theta(x_t,t,c)$ parameterized by $\theta$, This denoising model $\boldsymbol{\epsilon}_\theta(\cdot)$ is implemented with U-Net and trained with a mean-squared loss derived by a simplified variant of the variational bound:
$
\mathcal{L}=\mathbb{E}_{t,\boldsymbol{x}_0,\boldsymbol{\epsilon}}\left[\|\boldsymbol{\epsilon}-\hat{\boldsymbol{\epsilon}}_\theta(\mathbf{x}_t,t,\boldsymbol{c})\|^2\right]
$,
where $\boldsymbol{x}_0$ represents the real data with an additional condition $\boldsymbol{c}$, $t\in[0,T]$ denotes the time step of diffusion process, $\boldsymbol{x}_t=\alpha_t\boldsymbol{x}_{t-1}+\sigma_t\boldsymbol{\epsilon}$ is the noisy data at $t$ step, and $\alpha_t$, $\sigma_t$ are predefined functions of $t$ that determine the diffusion process.
% In Stable Diffusion, $\boldsymbol{c}$ generally represent the text features encoded from text prompts using CLIP \citep{CLIP}. In the denoising process, text features $\boldsymbol{c}$ are integrated into diffusion models through a cross-attention module. 
% 
% 
For conditional learning, classifier-free guidance \citep{Classifier-Free} is often employed, which uses a single neural network to parameterize both the conditional model and unconditional model, where for the unconditional model one can simply input a null token $\varnothing$ for the text features $c$ when predicting the noise, i.e. $\boldsymbol{\epsilon}_\theta(\boldsymbol{x}_t,t)=\boldsymbol{\epsilon}_\theta(\boldsymbol{x}_t,t,\boldsymbol{c}=\varnothing)$.
Then one can perform sampling using the following linear combination of the conditional
and unconditional noise estimates:
% \begin{equation}
$
\tilde{\boldsymbol{\epsilon}}_\theta(\boldsymbol{x}_t,t,\boldsymbol{c})=\omega\boldsymbol{\epsilon}_\theta(\boldsymbol{x}_t,t,\boldsymbol{c})+(1-\omega)\boldsymbol{\epsilon}_\theta(\boldsymbol{x}_t, t)$.
% \label{eq:conditional_learning}
% \end{equation}
Once the model $\boldsymbol{\epsilon}_\theta(\cdot)$ is trained, images can be generated from random noises in an iterative manner.

\textbf{IP-Adapter}  
As ``an image is worth a thousand words'', IP-Adapter \citep{ye2023ip} proposed an effective and lightweight adapter to achieve image prompt capability for the pre-trained text-to-image diffusion models.
It uses two decoupled cross-attention modules to process text and image conditions and finally performs linear weighting. 
Given the query features $\mathbf{Z}$, the text features $\boldsymbol{c}_t$, and the image features $\boldsymbol{c}_i$, the final formulation of the two cross-attention modules is defined as:
\begin{equation}
\begin{aligned}
\mathbf{Z}^{new}=\mathrm{Softmax}(\frac{\mathbf{Q}\mathbf{K}^{\top}}{\sqrt{d}})\mathbf{V}+\mathrm{Softmax}(\frac{\mathbf{Q}(\mathbf{K}^{\prime})^{\top}}{\sqrt{d}})\mathbf{V}^{\prime}\\
\mathrm{where~}\mathbf{Q}=\mathbf{Z}\mathbf{W}_{q},\mathbf{K}=\boldsymbol{c}_{t}\mathbf{W}_{k},\mathbf{V}=\boldsymbol{c}_{t}\mathbf{W}_{\upsilon},\mathbf{K}^{\prime}=\boldsymbol{c}_{i}\mathbf{W}_{k}^{\prime},\mathbf{V}^{\prime}=\boldsymbol{c}_{i}\mathbf{W}_{v}^{\prime}
\end{aligned}
\end{equation}
where $\mathbf{Q}$, $\mathbf{K}$ and $\mathbf{V}$ are the query, key, and values matrices from the text features;
$\mathbf{K}^{\prime}$ and $\mathbf{V}^{\prime}$ 
are the key, and values matrices from the image features. IP-Adapter used the same query for image cross-attention as for text cross-attention. The weight matrices $\mathbf{W}_q$, $\mathbf{W}_k$, and $\mathbf{W}_v$ correspond to text cross-attention and remain frozen, consistent with the original pre-trained model. Only the weight matrices in image cross-attention, $\mathbf{W}_k^{\prime}$ and $\mathbf{W}_v^{\prime}$, are trainable. 
In the inference stage, one can also adjust the weight of the image condition:
\begin{equation}
    \mathbf{Z}^{new}=\mathrm{Attention}(\mathbf{Q},\mathbf{K},\mathbf{V})+\lambda_{i}\cdot\mathrm{Attention}(\mathbf{Q},\mathbf{K}^{\prime},\mathbf{V}^{\prime})
    \label{eq:linear-weighting}
\end{equation}
where $\lambda_{i}$ is the coefficient of image conditions, and the model becomes the original text-to-image diffusion model if $\lambda_{i} = 0$.

% \begin{figure}
%   \centering
%   \includegraphics[width=1.0\linewidth]{figure/exp0.png}
%    \caption{\small{(a) The coefficient-tuning results of IP-Adapter \citep{ye2023ip}, InstantStyle \citep{wang2024instantstyle} and our proposed method. We use different coefficients for the image condition (i.e., $\lambda_{i}$ in Equation~\ref{eq:linear-weighting}). 
%    In the figure, satisfactory outcomes that effectively mitigate content from the reference while achieving style enhancement are highlighted with green boxes.
%    For IP-Adapter, lowering the weight factor for image condition helps to enhance the control ability of text prompts, but it also comes with style degradation. InstantStyle also struggles with maintaining object presence in text prompts at high values of $\lambda_i$. In contrast, our method delivers significantly more stable results across various $\lambda_i$ values.
%    (b) We report the mean opinion score \citep{huynh2010study} to assess the quality of the generated images. The significantly higher scores of our method across various $\lambda_i$ values demonstrate its robustness and effectiveness.
%    More comparison results are in Figure~\ref{fig: more-results-exp0} of Appendix~\ref{app-sec: more-results-exp0}.}}
% \label{fig:1}
% \end{figure}

\begin{figure}
  \centering
  \includegraphics[width=1.0\linewidth]{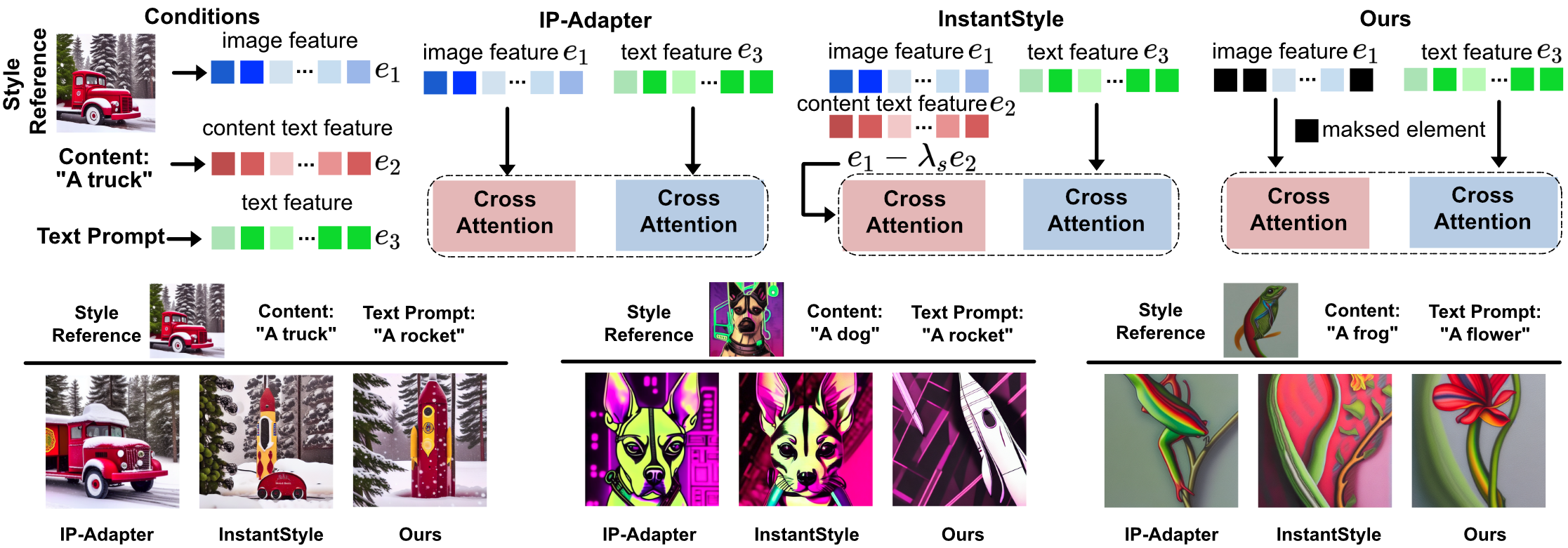}
  \vspace{-0.6cm}
   \caption{\textbf{Top:} The differences in the conditions between IP-Adapter \citep{ye2023ip}, InstantStyle \citep{wang2024instantstyle}, and Ours.
   We elaborate on how to select masked elements in Section~\ref{sec:Zero-Shot-LecDiff}.
   \textbf{Bottom:}  Illustration of the content leakages issue.}
\label{fig:overview}
\vspace{-0.6cm}
\end{figure}

\textbf{InstantStyle and Limitations}
Fully compatible with IP-Adapter, InstantStyle \citep{wang2024instantstyle} employs block-specific injection techniques to achieve style transfer. Additionally, it proposes an efficient method to decouple content and style from style references, highlighting that straightforward subtraction of content text features from image features can effectively reduce content leakages.
However, this approach has some limitations: 1) On the one hand, the feature subtraction is based on CLIP's embeddings, and it relies on the assumption that CLIP global features provide a robust characterization for explicit decoupling. It means that the process necessitates good alignment between CLIP's image-text features for all style references, which may be unrealistic for complex style references. 
2) On the other hand, tuning the coefficient of image condition (i.e., $\lambda_{i}$ in Equation~\ref{eq:linear-weighting}) is important for its effect in addressing content leakages, but it is labour-intensive and very tricky \footnote{We provide the generation results of InstantStyle across various coefficient values in Figure~\ref{fig: more-results-exp0} of Appendix~\ref{app-sec: more-results-exp0}, which showcasing that it heavily relies on test-time coefficient-tuning for style strength, requiring users to engage in a labour-intensive process to achieve a balanced synthesis between target content and style.}.

To comprehensively understand the content leakages issue, we present visualization examples for IP-Adapter and InstantStyle in the context of style transfer. Given a style-reference image as the additional image condition, the models are tasked with generating images that reflect the text prompt while incorporating the style of the image reference.
Based on Stable Diffusion V1.5 (SD1.5, \citep{rombach2022high}), we adopt DDIM sampler \citep{DDIM} with 30 steps, set the guidance scale to 7.5, and the coefficient of image condition to 1.0. 
As illustrated in Figure~\ref{fig:overview}, IP-Adapter struggles with maintaining the presence of objects in text prompts.
Due to image-text misalignment, InstantStyle also fails to achieve seamless synthesis of text prompts and reference styles.

\textbf{Effectiveness of Less Condition}
To mitigate image-text misalignment, we focus on feature manipulation within the image feature space instead of introducing feature subtraction between different modalities. 
% Based on the independence assumption among feature elements\ye{Why this assumption holds, it's not the key for our method}, 
Specifically, we propose to eliminate content from the style-reference image by discarding the image feature elements associated with that content. To achieve this, we mask the content-related elements by setting their values to zero. The content-related elements are identified through clustering the element-wise product of the style reference image features and the content text features.
% This zero-out operation follows the IP-Adapter's operation of zeroing out the CLIP image feature when the image condition is dropped. 
As shown in Figure~\ref{fig:overview}, our method successfully achieves more accurate style transfer by masking certain elements in the style reference. Moreover, as illustrated in Figure~\ref{fig: more-results-exp0},
compared to InstantStyle, our method can achieve more stable results across various coefficient values of $\lambda_i$, particularly in high-coefficient scenarios. 
Thus, we arrive at the key motivation of this paper:

{\textit{The experiment results in Figure~\ref{fig:1} and Figure~\ref{fig:overview} suggest that leveraging 
appropriately-selected fewer conditions, such as the masked image features, surprisingly effectively avoids content leakages, thereby enhancing text-to-image models in style transfer.}}
% \ye{suggest that fewer conditions can surprisingly effectively guide ...}

Motivated by this observation, we theoretically and experimentally explore the role of the masking strategy in eliminating content leakages in text-to-image diffusion models. In this paper, we demonstrate that our masking-based method can outperform recent state-of-the-art methods without the need for tuning model parameters or the coefficient value of $\lambda_{i}$, which we fix at 1.0.

\section{Methodology}
\label{Sec:Methodology}

In Section~\ref{sec:Zero-Shot-LecDiff}, we elaborate on our novel masking-based method for efficiently decoupling content from style. This approach utilizes a masking strategy for image features, where the masked {image feature} elements are identified through clustering on the element-wise product of image features and content text features. We also provide supporting evidence for the effectiveness of this masked element selection method in Proposition~\ref{proposition}.
Furthermore, in section~\ref{sec:superiority-of-less-condition}, we theoretically demonstrate that our method surpasses InstantStyle's feature subtraction by achieving a smaller divergence between the generated image distribution and the real image distribution (See Theorem~\ref{theorem:1}).
To delve deeper, we also investigate whether the effectiveness of appropriately fewer conditions holds in tuning-based models. We present the theoretical results in Theorem~\ref{theorem:2}.

\subsection{The Proposed Masking-Based Method for Decoupling Content form Style}
\label{sec:Zero-Shot-LecDiff}

Before delving into the details of our method, we first present the important notations as follows:

\begin{notations}
    Let $q(\boldsymbol{x}|c)$ be the joint distribution for the data $\boldsymbol{x}$ and the condition $c$ and $q(\boldsymbol{x}) = \sum_{c}q(\boldsymbol{x}|c)$.
    Let $p_{\boldsymbol{\theta, e}}(\boldsymbol{x})$ be a model parameterized by $\boldsymbol{\theta}\in\Theta$ and $\boldsymbol{e}\in {E}$, where $\boldsymbol{\theta}$ denotes the model parameters and $\boldsymbol{e}$ is the embedding for a condition. 
    In the context of style transfer, let $c_1$, $c_2$, and $c_3$ represent the style reference, the content text in style reference, and the target text prompt, respectively.
    \textcolor{black}{Models are tasked with generating a plausible target image by combining the content of $c_3$ with the style of $c_1$, while ensuring that the unwanted content $c_2$ does not transfer into the generated result.}
    The corresponding embeddings of $c_1$, $c_2$, and $c_3$ are denoted as $\boldsymbol{e}_1$, $\boldsymbol{e}_2$, and $\boldsymbol{e}_3$. Each of these embeddings is a $d$-dimensional feature. 
    Here, $\boldsymbol{e}_1$ is an image feature that is embedded by IP-Adapter's image encoder; $\boldsymbol{e}_2$ and $\boldsymbol{e}_3$ are text features embedded by the CLIP model.
    We denote the element-wise product feature between $\boldsymbol{e}_1$ and $\boldsymbol{e}_2$ as $\boldsymbol{e}_p$, i.e., $\boldsymbol{e}_p^i=\boldsymbol{e}_1^i \cdot \boldsymbol{e}_2^i$.
\end{notations}

% \textcolor{red}{\textbf{Task:} Given a style reference $c_1$, the content of the style reference $c_2$, and the target text prompt $c_3$, the text-driven style transfer aims to generate a plausible target image by combining the content of the target text prompt with the style of the style reference, while ensuring that the unwanted content $c_2$ from the style reference does not transfer into the generated result.}

\begin{figure}
  \centering
  \includegraphics[width=1.0\linewidth]{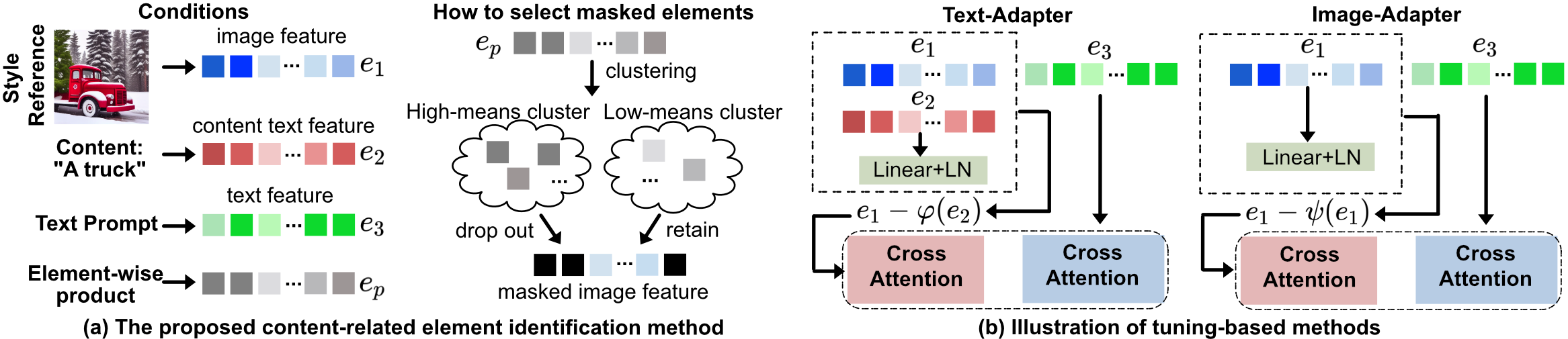}
  \vspace{-0.6cm}
   \caption{\small{
   (a) The proposed content-related elements identification method: we cluster the element-wise product between image and text features and directly discard elements in the high-means cluster;
   (b) Illustration of tuning-based models, \textcolor{black}{which we detail in Section~\ref{sec:superiority-of-less-condition}}. Text-Adapter and Image-Adapter learn the content feature from the text content feature and image feature, respectively. Only the newly added feature adapter modules (denoted as ``Linear+LN'') are trained while the pre-trained diffusion model is frozen.
   % We detail the Text-Adapter and Image-Adapter method in Section~\ref{sec:superiority-of-less-condition}.
   }}
   \vspace{-0.4cm}
   \label{fig:pipeline}
\end{figure}

To mitigate the image-text misalignment issue, we conduct feature manipulation within the latent space of image features.
We propose to eliminate content from the style reference by discarding the specific elements corresponding to that content.
To achieve this, we drop out these elements by setting their values to zero.
% \ye{\sout{following the IP-Adapter operation of zeroing out the CLIP image feature when the image condition is dropped.} Rephrase and emphasize our novelty}
As illustrated in Figure~\ref{fig:pipeline}~(a), the process unfolds as follows:

1) We first compute the element-wise products between the image feature $\boldsymbol{e}_1$ and the corresponding content text feature $\boldsymbol{e}_2$, denoting the results as $\boldsymbol{e}_p^i (i=1,\cdots,d)$ where $\boldsymbol{e}_p^i=\boldsymbol{e}_1^i \cdot \boldsymbol{e}_2^i$. 
2) Next, we cluster these elements $\boldsymbol{e}_p^i (i=1,\cdots,d)$ into $K$ classes.
3) We then generate a masking vector $\boldsymbol{m}$ based on the clustering results. 
For the element $\boldsymbol{e}_p^i$ in the highest-means cluster, \textcolor{black}{we set the corresponding element $\boldsymbol{m}^i$ to 0.}
4) Finally, we apply the mask vector $\boldsymbol{m}$ to the image feature $\boldsymbol{e}_1$ by computing $\boldsymbol{e}_1^\prime=\boldsymbol{e}_1 \odot \boldsymbol{m}$. This masked image feature $\boldsymbol{e}_1^\prime$, along with the text feature $\boldsymbol{e}_3$, is then incorporated into the cross-attention module of IP-Adapter.

The complete algorithm can be found in Algorithm~\ref{algorithm1} in Appendix~\ref{app-sec:algorithm}.
% Through Algorithm~\ref{algorithm1}, we explicitly remove the elements in the image features that correspond positively
% \ye{How to define correspond positively, correlation compute?} 
% with those in content text features. 
{With Algorithm~\ref{algorithm1}, we explicitly remove the image feature elements that contribute to the similarity between the image feature and content text feature. This means we identify and remove those elements that are most correlated with the content text feature in the feature space, thereby effectively reducing content leakages.}
Additionally, we provide a detailed explanation of why we utilize clustering on $\boldsymbol{e}_1^i \cdot \boldsymbol{e}_2^i$ to capture content-related elements, rather than relying on the absolute difference between $\boldsymbol{e}_1^i$ and $ \boldsymbol{e}_2^i$ or other metrics. 
Through theoretical exploration, we discovered that masking strategy with on $\boldsymbol{e}_1^i \cdot \boldsymbol{e}_2^i$ leads to the highest energy score for content {text feature} $\boldsymbol{e}_2$, 
% \ye{feature elements}, 
{thus} effectively decoupling content from style. We prove this advantage in Proposition~\ref{proposition}). 
Before the proposition, we first give a brief introduction to Energy Diffusion Guidance as follows:

\textbf{Energy Diffusion Guidance} 
Given the noisy image $\mathbf{x}_t$ and the domain of the given conditions $c$, Yu et al. \citep{yu2023freedom} proposed a energy diffusion guidance to model the gradient $\nabla_{\mathbf{x}_t}\log p(c|\mathbf{x}_t)$ by resorting to the energy function:
$p(c|\mathbf{x}_t)=\frac{\exp(-\lambda\mathcal{E}(c, \mathbf{x}_t))}{Z}$,
where $\lambda$ denotes the positive temperature coefficient and $Z>0$ denotes the normalizing constant, computed as $Z=\int_{c\in\mathcal{C}}\exp\{-\lambda\mathcal{E}(c,\mathbf{x}_t)\}$, $\mathcal{E}(c,\mathbf{x}_t)$ is an energy function that measures the compatibility between the condition $c$ and the noisy image $\mathbf{x_t}$. The value of the energy function will be smaller when $c$ is more compatible with $\mathbf{x_t}$.
Therefore, the gradient $\nabla_{\mathbf{x}_t}\log p(c|\mathbf{x}_t)$ can be implemented with the following:
$\nabla_{\mathbf{x}_t}\log p(c|\mathbf{x}_t)\propto-\nabla_{\mathbf{x}_t}\mathcal{E}(c,\mathbf{x}_t)$, which is referenced to the \textit{energy guidance}.
% We use the time-independent distance measuring functions $\mathcal{E}(c,\mathbf{x}_0)$ to approximate the energy function as:
Following \citep{yu2023freedom}, we use a time-independent distance measuring functions $\mathcal{D}_{\boldsymbol{\theta}}(c,\mathbf{x}_0)$  to approximate the energy function $\mathcal{E}(c,\mathbf{x}_0)$:
\begin{equation}
    \mathcal{E}(c,\mathbf{x}_t)\approx\mathbb{E}_{p(\mathbf{x}_0|\mathbf{x}_t)}\mathcal{D}_{\boldsymbol{\theta}}(c,\mathbf{x}_0)
    \label{eq:energy}
\end{equation}
where $\boldsymbol{\theta}$ defines the model parameters of the diffusion model. $\mathcal{D}_{\boldsymbol{\theta}}(c,\mathbf{x}_0)$ computes the cosine similarity between the CLIP embeddings of the given condition $c$ and image $\mathbf{x}_0$. 
% The distance measuring function can be calculated as $\mathbb{E}_{p(\mathbf{x}_0|\mathbf{x}_t)}\mathcal{D}_{\boldsymbol{\theta}}(c,\mathbf{x}_0)= Dist(\mathcal{P}_{\boldsymbol{\theta}_1}(c),\mathcal{P}_{\boldsymbol{\theta}_2}(\mathbf{x}_{0|t}))$.
% Here, $Dist(\cdot)$ denotes the distance measuring methods like
% Euclidean distance, $\mathcal{P}_{\boldsymbol{\theta}_1}(\cdot)$ and 
% $\mathcal{P}_{\boldsymbol{\theta}_2}(\cdot)$
% project the condition and image to the same space for distance measurement.

\begin{proposition}
{\textbf{[The superiority of the proposed masked element selection method]}}
We denote the masked elements in the image feature as $\boldsymbol{e}_1^{s+1}, \cdots, \boldsymbol{e}_1^{d}$ and denote the feature composed by these elements as $\boldsymbol{e}_1^m$, i.e., $\boldsymbol{e}_1^m :=[\boldsymbol{e}_1^{s+1}, \cdots, \boldsymbol{e}_1^{d}]$.
Incorporating the masking strategy, i.e., $\boldsymbol{e}_1^m=\varnothing$ leads to $\nabla_{\mathbf{x}_t}\log p(\boldsymbol{e}_1^m|\mathbf{x}_t,\boldsymbol{e}_3)=0$. 
According to $\nabla_{\mathbf{x}_t}\log p(c|\mathbf{x}_t)\propto-\nabla_{\mathbf{x}_t}\mathcal{E}(c,\mathbf{x}_t)$, 
% energy function $\mathcal{E}(c,\mathbf{x}_t)$ arrives at a local maximization, which means 
we have approximated the local maximization of 
% $\mathcal{E}(\boldsymbol{e}_1^{m},\mathbf{x}_t)$ 
$\mathcal{D}_{\boldsymbol{\theta}}(\boldsymbol{e}_1^{m},\mathbf{x}_{0|t})$
since $\mathcal{E}(\boldsymbol{e}_1^{m},\mathbf{x}_t)
    \approx \mathcal{D}_{\boldsymbol{\theta}}(\boldsymbol{e}_1^{m},\mathbf{x}_{0|t})$.
The proposed masking strategy enforces the selected component $\boldsymbol{e}_1^m$ to have the closest distance with the content text feature $\boldsymbol{e}_2$.
Therefore, according to the relation between energy and distance as defined in Equation~\ref{eq:energy}, the masking strategy with clustering on $\boldsymbol{e}_1^i \cdot \boldsymbol{e}_2^i$ can lead to the highest energy score for the content text feature $\boldsymbol{e}_2$ when compared to other masking methods.
\label{proposition}
\vspace{-0.2cm}
\end{proposition}

This proposition indicates that our proposed masking-based method not only mitigates the image-text misalignment issue by manipulating embeddings within the image feature space, but also achieves the highest energy score for content text feature compared to other masked element selection methods. This effectively reduces the likelihood of content text features, leading to superior performance in content removal.

\subsection{Theoretical Evidence of Fewer Conditions in Enhancing Style Transfer}
\label{sec:superiority-of-less-condition}

In this section, we present the theoretical evidence supporting our method's superiority over InstantStyle and IP-Adapter. Theorem~\ref{theorem:1} illustrates the advantages of our masking-based approach, suggesting that fewer conditions can achieve a smaller divergence between the generated image distribution and the real image distribution. To delve deeper, we also investigate whether a tuning-based model guided by fewer conditions can yield improved results.  Our findings indicate that training models with fewer conditions can also enhance style transfer, as illustrated in Theorem~\ref{theorem:2}.

\begin{notations}
    We use $p_{\boldsymbol{\theta, e}}(\boldsymbol{x})$ to approximate the conditional data distribution $q(\boldsymbol{x}|c)$. Let $p_{\boldsymbol{\theta}, \boldsymbol{\phi}}(\boldsymbol{x}|c) = p_{\boldsymbol{\theta}, \boldsymbol{e}}(\boldsymbol{x})|_{\boldsymbol{e}=\boldsymbol{\phi}(c)}$, where $\boldsymbol{e}$ denotes the embedding of the given condition $c$.
We denote a certain statistics divergence as $\mathcal{D}$ (or more loosely a divergence upper bound). 
\end{notations}

\begin{theorem}
\textbf{[Why the masking strategy is better]}
Suppose the divergence $\mathcal{D}$ is convex, and the elements in the image feature are independent of each other. 
% We denote the element in the image feature and text feature as $\boldsymbol{e}_1^{i} (i\in\{1, \cdots, d\})$ and $\boldsymbol{e}_2^{i} (i\in\{1, \cdots, d\})$, respectively.
We denote $\boldsymbol{e}_1:=\boldsymbol{e}_1^{i,\cdots,d}$, $\boldsymbol{e}_2:=\boldsymbol{e}_2^{i,\cdots,d}$, and $\boldsymbol{e}_3:=\boldsymbol{e}_3^{i,\cdots,d}$.
Thus, the divergence between the generated and ground-truth image distribution of the InstantStyle model is:
$$
D_1 = \mathbb{E}_{q(\boldsymbol{e}_1^{i, \cdots, d}, \boldsymbol{e}_2^{i, \cdots, d}, \boldsymbol{e}_3^{i, \cdots, d})}\mathcal{D}(q(\boldsymbol{x}|\boldsymbol{e}_1, \boldsymbol{e}_2, \boldsymbol{e}_3)\|p_{\boldsymbol{\theta}}(\boldsymbol{x}|\boldsymbol{e}_1^{i, \cdots, d}, \boldsymbol{e}_2^{i, \cdots, d}, \boldsymbol{e}_3^{i, \cdots, d}))
$$
We denote the masked element in the image feature as $\boldsymbol{e}_1^{s+1}, \cdots, \boldsymbol{e}_1^d$.
Thus, the divergence result of the proposed masking strategy is:
$$
D_2 = \mathbb{E}_{q(\boldsymbol{e}_1^{i, \cdots, d}, \boldsymbol{e}_2^{i, \cdots, d}, \boldsymbol{e}_3^{i, \cdots, d})}\mathcal{D}(q(\boldsymbol{x}|\boldsymbol{e}_1, \boldsymbol{e}_2, \boldsymbol{e}_3)\|p_{\boldsymbol{\theta}}(\boldsymbol{x}|\boldsymbol{e}_1^{i, \cdots, s}, \boldsymbol{e}_2^{i, \cdots, d}, \boldsymbol{e}_3^{i, \cdots, d}))
$$
With the assumption:
$$\mathbb{E}_{q(\boldsymbol{e}_1^{i, \cdots, d})} \mathcal{D}(q(\boldsymbol{x}|\boldsymbol{e}_1^{i, \cdots, d})\| p_{\theta}(\boldsymbol{x}|\boldsymbol{e}_1^{i, \cdots, s})) \leq  \mathbb{E}_{q(\boldsymbol{e}_1^{i, \cdots, s})} \mathcal{D}(q(\boldsymbol{x}|\boldsymbol{e}_1^{i, \cdots, s})\| p_{\theta}(\boldsymbol{x}|\boldsymbol{e}_1^{i, \cdots, s}))$$ and by Jensen's inequality, we have $D_2 \leq D_1$.
\label{theorem:1}
\end{theorem}

This theorem indicates that: {compared to InstantStyle, 
masking certain elements (i.e., $\boldsymbol{e}_1^{s+1}, \cdots, \boldsymbol{e}_1^d$) of the image feature} achieves a smaller divergence between the generated and ground-truth image distribution. 
{Further, we also investigate whether a tuning-based model conditioned on appropriately fewer conditions can yield improved results.} 
We begin by formalizing the learning paradigms of tuning-based models, \textcolor{black}{as illustrated in Figure~\ref{fig:pipeline}~(b)}, as follows:

\textbf{Learning Paradigms}
\textit{Text-Adapter:} Given the style-reference image $c_1$, the content in style reference $c_2$, and the target text prompt $c_3$, models are tasked with generating images that reflect the text prompt $c_3$ while incorporating the style of $c_1$ and avoiding the presence of content $c_2$.
Without loss of generality, we keep the parameters in condition encoders frozen and perform adapter tuning for content text feature $\boldsymbol{e}_2$ with adapter $\boldsymbol{\phi}$. 
\textcolor{black}{The style reference's image feature $e_1$ is subtracted by the content feature $\phi(e_2)$ to avoid the presence of content $c_2$.}
The text adapter $\boldsymbol{\phi}$ can be optimized by:
$$
   \min_{\boldsymbol{\phi}} \mathbb{E}_{q(c_1, c_2,  c_3)}\mathcal{D}(q(\boldsymbol{x}|c_1, c_2,  c_3)\|p_{\boldsymbol{\theta},\boldsymbol{\phi}}(\boldsymbol{x}|c_1, c_2,  c_3))
$$
\textit{Image-Adapter:} Instead of tuning the content text feature, we directly extract the content feature from the style-reference image. 
We denote the extraction function (image adapter) as $\boldsymbol{\psi}$. Then we have the final image feature incorporated into the cross-attention module represented as $\boldsymbol{e}_1 - \boldsymbol{\psi}(\boldsymbol{e}_1)$.
Thus, the optimization objective is:
$$
    \min_{\boldsymbol{\psi}} \mathbb{E}_{q(c_1, c_2,  c_3)}\mathcal{D}(q(\boldsymbol{x}|c_1, c_2,  c_3)\|p_{\boldsymbol{\theta}, \boldsymbol{\psi}}(\boldsymbol{x}|c_1, c_3))
$$
\textcolor{black}{
% As illustrated in Figure~\ref{fig:pipeline}~(b),
% the style reference's image feature $e_1$ is subtracted by the content feature $\phi(e_2)$ or $\psi(e_1)$ to avoid the presence of content $c_2$.
In practice, the feature adapters (denoted as ``Liner+LN'') in Figure~\ref{fig:pipeline}~(b) are trained through image reconstruction using mean squared error (MSE) loss to predict reconstruction errors. The specific algorithm is provided in Algorithm~\ref{algoritm2} in Appendix~\ref{app-sec:algorithm}. We instantiate the two optimization objectives, as shown in the 9th line of Algorithm~\ref{algoritm2}, for the Text-Adapter and Image-Adapter, respectively.
% We highlight the optimization objective in Algorithm~\ref{algoritm2} for the Text-Adapter and Image-Adapter, respectively. These optimization objectives are used to minimize the prediction error of noise in reconstructing style reference while maximizing the difference between the model conditioned on the style reference's content and that conditioned on the target prompt.
}

\vspace{-0.4cm}
\begin{assumption}
Suppose the condition $c_1 \in \mathcal{C}_1$ and $c_3 \in \mathcal{C}_3$ are independent of each other, and the condition $c_2$ is dependent on the style reference $c_1$. 
Given the condition $c_1=x$, the content information $c_2=y$ is uniquely determined. 
% Thus, we have $q(c_2)q(c_1,c_3) \geq q(c_1, c_2, c_3)$, since $q(c_1=x, c_2=y, c_3=z)=q(c_1=x,c_3=z)$ and $q(c_1=x, c_2\neq y, c_3=z)=0$.
% By the Tower Law and non-negativity of $\mathcal{D}$, we have 
% $\mathbb{E}_{q(c_1, c_2,  c_3)} \mathcal{D}(q(\boldsymbol{x}|c_1, c_2,  c_3)\| p_{\theta, \boldsymbol{\psi}}(\boldsymbol{x}| c_1, c_3)) 
% \leq
% \mathbb{E}_{q(c_2|c_1, c_3)}\mathbb{E}_{q(c_1, c_3)} \mathcal{D}(q(\boldsymbol{x}|c_1, c_2,  c_3)\| p_{\theta, \boldsymbol{\psi}}(\boldsymbol{x}| c_1, c_3)) = 
% \mathbb{E}_{q(c_1, c_3)}
% \mathcal{D}(q(\boldsymbol{x}|c_1, c_3)\| p_{\theta, \boldsymbol{\psi}}(\boldsymbol{x} |c_1, c_3))$. 
% It is straightforward to extend this to the minimization case, leading to the following inequality:
% $$\min_{\boldsymbol{\psi}} \mathbb{E}_{q(c_1, c_2,  c_3)}\mathcal{D}(q(\boldsymbol{x}|c_1, c_2,  c_3)\|p_{\boldsymbol{\theta},\boldsymbol{\psi}}(\boldsymbol{x}|c_1, c_3)) \leq 
%     \min_{\boldsymbol{\psi}} \mathbb{E}_{q(c_1, c_3)}\mathcal{D}(q(\boldsymbol{x}|c_1, c_3)\|p_{\boldsymbol{\theta},\boldsymbol{\psi}}(\boldsymbol{x}|c_1, c_3)) $$
\label{assumption}
\end{assumption}
% \vspace{-0.4cm}
Under Assumption~\ref{assumption}, we reveal that a tuning-based model conditioned on appropriately fewer conditions can yield improved style transfer results:

\begin{theorem}{\textbf{[The superiority of tuning-based model conditioned on fewer conditions]}}
Suppose the divergence $\mathcal{D}$ is convex, and the function space $\Phi$ and $\Psi$ ($\boldsymbol{\phi} \in \Phi$ and $\boldsymbol{\psi} \in \Psi$) includes all measurable functions. 
Under Assumption~\ref{assumption}
and by Jensen's inequality, we have:
% \begin{equation}
% \begin{aligned}
%     \mathcal{D}(q(\boldsymbol{x}|c_1, c_3)\|p_{\boldsymbol{\theta},\boldsymbol{\psi}}(\boldsymbol{x}|c_1, c_3)) 
% & = \mathcal{D}(\mathbb{E}_{q(c_2)}q(\boldsymbol{x}|c_1, c_2,  c_3)\|\mathbb{E}_{q(c_2)}p_{\boldsymbol{\theta}, \boldsymbol{\psi}}(\boldsymbol{x}|c_1, c_2,  c_3))\\ 
% & \leq \mathbb{E}_{q(c_2)}\mathcal{D}(q(\boldsymbol{x}|c_1, c_2,  c_3)\|p_{\boldsymbol{\theta},\boldsymbol{\psi}}(\boldsymbol{x}|c_1, c_2,  c_3))
% \end{aligned}
%  \label{eq:2}
% \end{equation} 
% Combining Equation~\ref{eq:1} and Equation~\ref{eq:2}, we have:
\begin{equation}
\begin{aligned}
    \min_{\boldsymbol{\psi}} \mathbb{E}_{q(c_1, c_2,  c_3)}\mathcal{D}
    & (q(\boldsymbol{x}|c_1, c_2,  c_3)\|p_{\boldsymbol{\theta},\boldsymbol{\psi}}(\boldsymbol{x}|c_1, c_3)) \\
& \leq  
    \min_{\boldsymbol{\phi}} \mathbb{E}_{q(c_1, c_2,  c_3)}\mathcal{D}(q(\boldsymbol{x}|c_1, c_2,  c_3)\|p_{\boldsymbol{\theta},\boldsymbol{\phi}}(\boldsymbol{x}|c_1, c_2,  c_3)) 
\end{aligned}
\end{equation}
\label{theorem:2}
\vspace{-0.7cm}
\end{theorem}

This theorem indicates that learning content features based on the text feature 
results in a larger divergence between the generated and real image distribution, compared to learning content features directly within the feature space of image features. Overall, these theoretical results demonstrate that appropriately fewer conditions boost better text-to-image diffusion models in style transfer.

% \vspace{-0.2cm}
\section{Experiments}
\label{sec:experiments}

In this section, we first demonstrate the proposed theoretical results.
Previous evaluation datasets do not contain explicitly defined references' contents, thus making it inaccurate in evaluating content leakages. Instead, we consider an evaluation dataset comprising various defined reference contents and styles for comprehensively assessing models' capability in addressing content leakages.
In this paper, we construct the evaluation dataset consisting of 10 content objects and 21 image styles.
Extensive experimental results demonstrate that both tuning-free and tuning-based models, conditioned on appropriately selected fewer conditions, achieve higher text fidelity and style similarity, which aligns well with Theorem~\ref{theorem:1}-\ref{theorem:2}.
Next, we report our method's performance across various styles and compare it with existing approaches using the StyleBench \citep{gao2024styleshot} benchmark. Experimental results demonstrate the proposed method's effectiveness in avoiding content leakages.
% surpassing recent state-of-the-art approaches in achieving a seamless balance between content preservation and style enhancement.

\subsection{Quantitative Analysis of Our Method in Addressing Content Leakages}
\label{sec:exp1}

\textbf{Evaluation Dataset}
We construct the evaluation dataset using the 10 classes of CIFAR-10 \citep{krizhevsky2009learning}. 
Leveraging the code of MACE \citep{lu2024mace}, we generate 21 distinct styles for each class, each containing 8 variations. The dataset is divided into two subsets based on image style for training and testing. Using these generated images as references, we train tuning-based models (i.e., Image-Adapter and Text-Adapter) through image reconstruction. During inference, we utilize the test dataset as image references to conduct text-driven style transfer for 5 text prompts. Additional details about the datasets are provided in Table~\ref{tab:datasets-in-exp1} of Appendix~\ref{app-sec: datasets-in-exp1}.

\textbf{Model Configuration} For the tuning-based models, we update adapter weights for 2500 steps using Adam optimizer \citep{kingma2014adam} with a learning rate of 0.00001. We adopt the same adapter layer structure for Image-Adapter and Text-Adapter, which consists of a linear layer and a batch normalization.
% We train models by image reconstruction using mean square error (MSE) loss for error predictions. The specific algorithm is in Algorithm~\ref{algoritm2} of Appendix~\ref{app-sec:algorithm}.
Subsequently, leveraging the test data as image references, the trained model is utilized to generate stylized images for 5 text prompts. 
For all experiments, we adopt Stable Diffusion V1.5 as our base text-to-image model, and we set the clustering classes to 2 for our masking-based method.

\begin{table}[H]
\centering
\vspace{-0.3cm}
\caption{Quantitative comparison with advanced text-driven style transfer methods.
We mark the \textbf{Best results} and underscore the \underline{second best results}.}
\vspace{0.1cm}
\resizebox*{1.0\linewidth}{!}{
\begin{tabular}{llllllll} 
\hline
 Method   & StyleCrafter & StyleAligned & StyleDrop & DEADiff  & InstantStyle & StyleShot & Ours \\ 
\hline
% style score $\uparrow$    & 0.290        & 0.388        & 0.278    & 0.252       & 0.456    & 0.279    & 0.301    & 0.267    \\
% fidelity score $\uparrow$    & 0.240        &  0.213        & 0.250    & 0.250      & 0.171    & 0.255     & 0.227    & 0.256    \\
% leakage score  $\downarrow$ & 0.142      & 0.173      & 0.149    & 0.127      & 0.218     & 0.136     & 0.135    & 0.120   \\
% \hline
style score $\uparrow$    & 0.245    &  0.244     & 0.240   &    0.230   & \textbf{0.290}    & {0.267}   & \underline{0.273}       \\
fidelity score $\uparrow$    & 0.858    & 0.662     & 0.889    & 0.916      & 0.820     & \underline{0.956}   & \textbf{0.972}       \\
leakage score  $\downarrow$ & 0.589    & 0.720      & 0.600   & \underline{0.523}      & 0.596    & 0.543   &  \textbf{0.478}       \\
Human Preference $\uparrow$   &  1.7\%     & 4.7\%   & 19.6\%    & 4.7\%       & 8.3 \%    &  \underline{24.3\%}        & \textbf{36.7\%}       \\
\hline
\end{tabular}
}
\label{tab:exp1-overall}
\vspace{-0.5cm}
\end{table}

\textbf{Evaluation Metrics}
\textit{Human Preference}: Following \citep{liu2023stylecrafter, qi2024deadiff}, we conduct user preference studies to evaluate models' style transfer ability.
Compared to other methods, our method achieves the highest human preferences by a large margin, demonstrating robust stylization across various styles and responsiveness to text prompts.
\textit{CLIP-Based Scores}: We also assess the quality of generated images using CLIP \citep{CLIP} with ViT-H/14 as the image encoder. 
% We present these evaluation results for reference purposes only. We compute the cosine similarity of features between the generated images and the text prompt, referred to as the \textbf{fidelity score $\uparrow$}; the similarity between the generated images and the references' content text, termed the \textbf{leakage score $\downarrow$}; and the similarity between the generated images and the style reference, adjusted by subtracting the content score, referred to as the \textbf{style score $\uparrow$}.  The fidelity score measures text fidelity, the style score assesses style similarity with the style reference, and the leakage score indicates content leakages from the style reference, where a lower score is preferable.
\textcolor{black}{We perform binary classification using the CLIP model \cite{CLIP} 
% on the generated images to differentiate between the reference's content text and the text prompt. 
on the generated images to distinguish between the reference's content text and the text prompt.}
% \textcolor{red}{Here, the binary classification is performed using the CLIP model with the CLIP-H/14 as the image encoder.}
The computed classification accuracy is referred to as the \textbf{fidelity score $\uparrow$}. We also calculate the similarity between the generated images and the reference's content text, calibrated by the similarity between the reference images and the content text, termed the \textbf{leakage score $\downarrow$}. Finally, we assess the similarity between the generated images and the style reference, adjusted by subtracting the leakage score, which we refer to as the \textbf{style score $\uparrow$}. 
The fidelity score measures the fidelity to the text instructions, the style score assesses style similarity with the style reference, and the leakage score indicates content leakages from the style reference, where a lower score is preferable. 
More details of these scores are provided in Appendix~\ref{app-sec: datasets-in-exp1}.

\begin{figure}[H]
  \centering
  \includegraphics[width=1.0\linewidth]{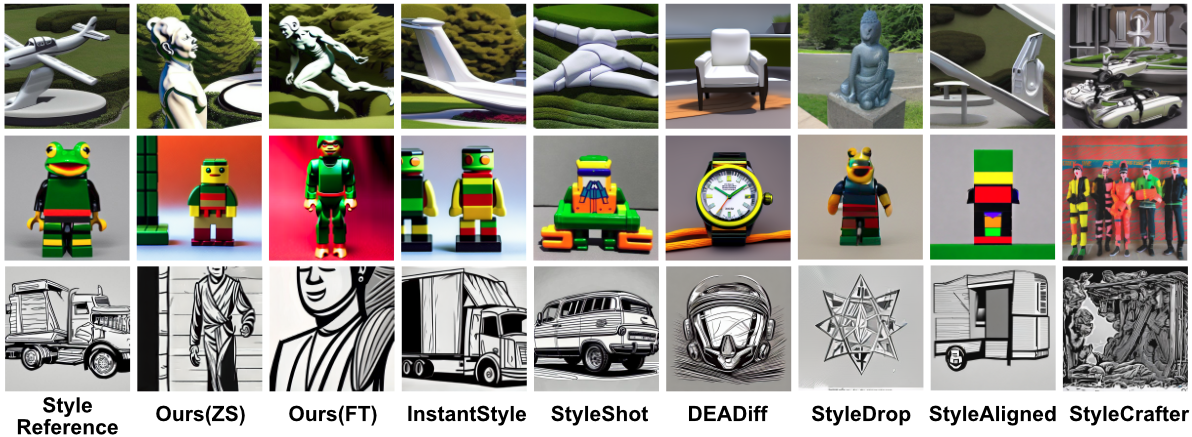}
  \vspace{-0.7cm}
   \caption{
   In the figure, \textcolor{blue}{the text prompt is ``A human''.} 
   Leveraging appropriately fewer conditions, Ours(ZS) and Ours(FT) denote the proposed masking-based method and the tuning-based Image-Adapter method, respectively. \textbf{Our methods successfully transfer the references' styles without content leakages.} More results can be found in Figure~\ref{fig:more-results-exp1-1} and Figure~\ref{fig:more-results-exp1-2}-\ref{fig:more-results-exp1-6} in Appendix~\ref{app-sec: more-results-exp1}.}
\label{fig:4}
\vspace{-0.6cm}
\end{figure}

\textbf{Experiment Results}
We provide visual comparisons between the proposed masking-based method and state-of-the-art methods in Figure~\ref{fig:4}, 
demonstrating that our approach can mitigate content leakages without introducing style degradations. 
These results align well with the theoretical findings of Theorem~\ref{theorem:1}, showcasing that fewer conditions more effectively address both content leakages and style degradations.

\vspace{-0.2cm}

\vspace{-0.2cm}
\begin{figure}[H]
  \centering
  \includegraphics[width=1.0\linewidth]{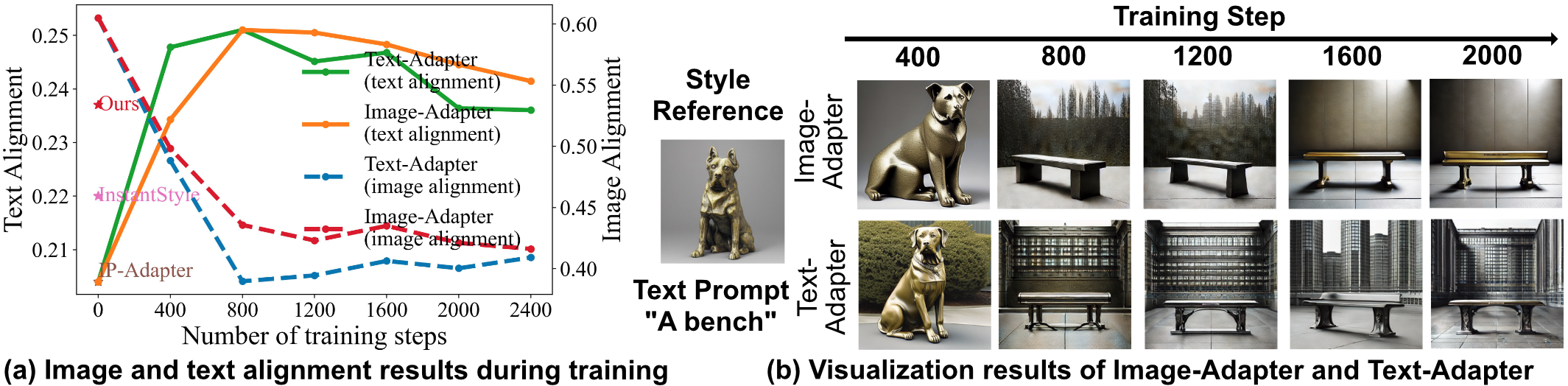}
  \vspace{-0.65cm}
   \caption{
   Comparison between the Image-Adapter and Text-Adapter model. 
   (a) Following \citep{gao2024styleshot}, we report the image and text alignment scores alongside training steps. 
   % {Image alignment refers to the cosine similarity between the generated images and the CLIP features of the style reference, while text alignment measures the cosine similarity between the generated images and the CLIP features of the text prompt.}
   We also present the tuning-free models' 
   (i.e., IP-Adapter, InstantStyle, and our masking-based method) 
   fidelity scores in the figure. 
   (b) Visual comparisons between Image-Adapter and Text-Adapter. 
   }
\label{fig:tuning-based-model-results}
\vspace{-0.6cm}
\end{figure}

For the tuning-based models, we present their image and text alignment scores and generation results along with the training steps in Figure~\ref{fig:tuning-based-model-results}. The observations are as follows:
1) Guided by appropriately-selected fewer conditions, the Image-Adapter model outperforms the Text-Adapter in both style scores and fidelity scores, indicating a smaller distribution divergence between the generated images and real images, consistent with the theoretical results of Theorem~\ref{theorem:2}.
2) As shown in Figure~\ref{fig:tuning-based-model-results}~(b), while the Text-Adapter reduces content leakages, it leads to significant style degradation as the training steps increase. In contrast, by leveraging fewer conditions, the Image-Adapter successfully avoids the image and text modal misalignment, with no content leakages while achieving style enhancement.

Overall, we provide quantitative comparisons with recent methods, such as StyleCrafter \citep{liu2023stylecrafter}, StyleAligned \citep{StyleAligned}, StyleDrop \citep{sohn2024styledrop}, DEADiff \citep{qi2024deadiff}, StyleShot~\citep{gao2024styleshot} and so on.
We present the CLIP-based score results in Table~\ref{tab:exp1-overall}.
Additional visual comparisons can be found in Figure~\ref{fig:more-results-exp1-1} and  Figure~\ref{fig:more-results-exp1-2}-\ref{fig:more-results-exp1-6} of Appendix~\ref{app-sec: more-results-exp1}. The overall best performances in CLIP-based scores and human preferences further demonstrate the effectiveness of our method in balancing content leakage mitigation with style enhancement.

% \begin{table}[H]
% \centering
% \vspace{-0.3cm}
% \caption{Quantitative comparison with advanced text-driven style transfer methods. We mark the \textbf{Best results} and \underline{second best results}. Ours(ZS) and Ours(FT) denote the proposed tuning-free and the tuning-based Image-Adapter method, respectively.}
% \vspace{0.1cm}
% \resizebox*{1.0\linewidth}{!}{
% \begin{tabular}{lllllllll} 
% \hline
%   CLIP       & StyleCrafter & StyleAligned & StyleDrop & DEADiff  & InstantStyle & StyleShot & Ours(ZS) & Ours(FT)  \\ 
% \hline
% image score $\uparrow$    & 0.452        & \underline{0.520}        & 0.454     & 0.434        & \textbf{0.556}        & 0.451     & 0.475    & 0.454     \\
% text score $\uparrow$    & 0.253        & 0.247        & \textbf{0.263}     & \underline{0.262}    & 0.204            & \textbf{0.263}     & 0.237    & \textbf{0.263}     \\
% content score  $\downarrow$ & 0.206        & 0.225        & 0.207     & \textbf{0.198}       & 0.246        & \underline{0.201}     & 0.209    & {0.206}     \\
% \hline
% Human      & StyleCrafter & StyleAligned & StyleDrop & DEADiff  & InstantStyle & StyleShot & Ours(ZS) & Ours(FT)  \\ 
% \hline
% image $\uparrow$   & 4.0\%        & 6.0\%        &  10.0\%    & 14.0\%    &  4.0\%     &  \underline{16.7\%}          & \underline{16.7\%}     &  \textbf{29.3\%}     \\
% text $\uparrow$    & 1.3\%       &  1.3\%       & 8.0\%    & 4.7\%     & 4.0\%       & 10.0\%     & \underline{20.0\%}    &   \textbf{49.3\%}  \\
% \hline
% \end{tabular}
% }
% \label{tab:exp1-overall}
% \vspace{-0.3cm}
% \end{table}

\subsection{Comparison with State-of-the-art Methods on StyleBench}

\vspace{-0.3cm}
\begin{figure}[H]
  \centering
  \includegraphics[width=1\linewidth]{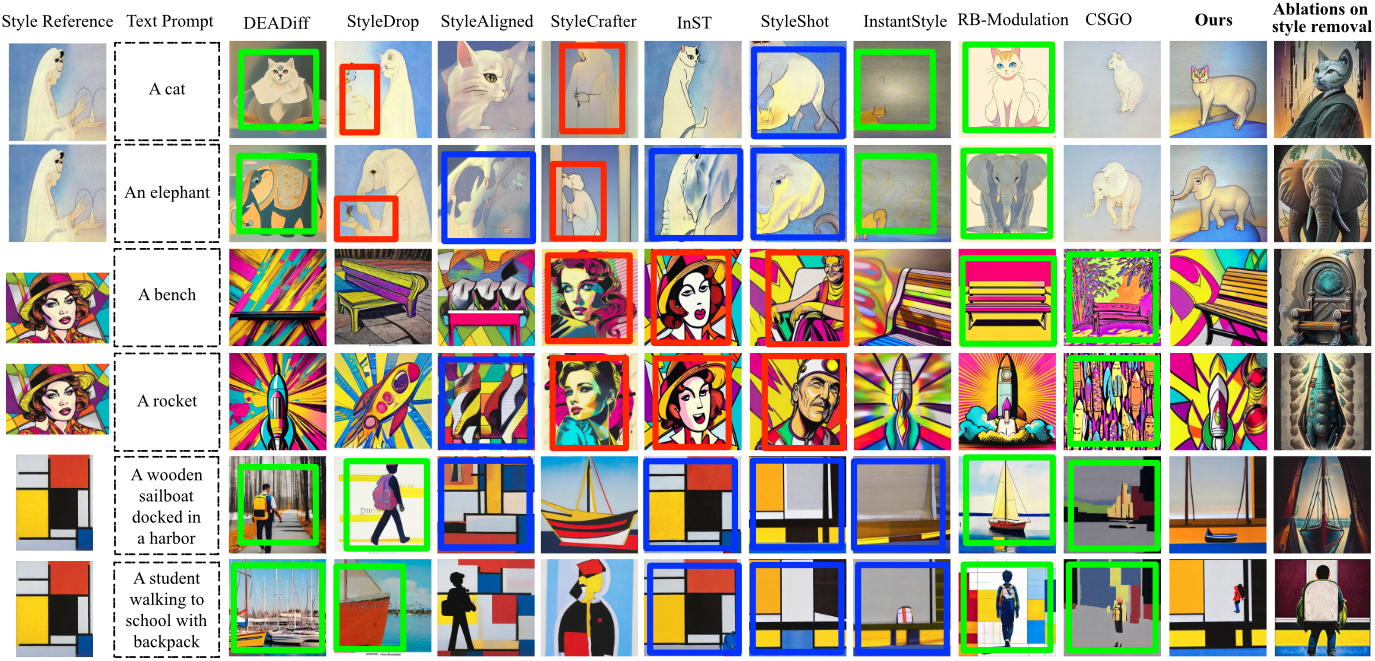}
  \vspace{-0.68cm}
   \caption{Visual comparison between recent state-of-the-art methods, including \textcolor{black}{RB-Modulation \citep{rout2024rb}, CSGO \citep{xing2024csgo}, for addressing content leakage,} and ours in text-driven style transfer. We mark the results with significant \textcolor{red}{content leakages}, \textcolor{green}{style degradation}, and \textcolor{blue}{loss of text fidelity} with \textcolor{red}{red}, \textcolor{green}{green}, and \textcolor{blue}{blue} boxes, respectively.
   The proposed masking-based method does not require content knowledge of the image reference; instead, we leverage the CLIP text feature of ``person, animal, plant, or object in the foreground'' to identify the elements that need to be masked.
   % The ablation study results in the last column demonstrate that our masking-based method can efficiently and accurately decouple content from style. 
   \textcolor{black}{More comparison examples are in Figure~\ref{fig:more-results-on-StyleBench} of Appendix~\ref{app-sec: more-results-exp1}.}
   }
\vspace{-0.5cm}
\label{fig:5}
\end{figure}

% \textbf{Evaluation Benchmark}
\textbf{Experiment Details}
To comprehensively evaluate the effectiveness of the proposed masking-based method, we conduct evaluations on the recent style transfer benchmark StyleBench \citep{gao2024styleshot},
% both on the text-driven style transfer and image-driven style transfer. 
which covers 73 distinct styles, ranging from paintings, flat illustrations to sculptures with varying materials. 
% We adopt Stable Diffusion V1.5 as our base text-to-image model.
For InstantStyle and our method, we employ the feature subtraction and masking strategy, respectively, on the extracted image features by StyleShot.

\begin{table}
\centering
\caption{Quantitative comparison with state-of-the-art text-driven style transfer methods. 
% As mentioned in previous studies \citep{sohn2024styledrop}, text and image alignment scores are not ideal for evaluation in style transfer tasks. We present the evaluation results in this paper only for reference. 
The highest human preference scores demonstrate the effectiveness of our method.
}
\vspace{0.1cm}
\resizebox*{1.0\linewidth}{!}{
\begin{tabular}{lllllllll} 
\hline
Method  & StyleCrafter & DEADiff & StyleDrop & InST   & StyleAligned & StyleShot & InstantStyle & Ours  \\ 
\hline
text alignment $\uparrow$  & 0.202        & 0.232    & 0.220     & 0.204  & 0.213        & 0.219     & 0.275            & 0.265      \\
image alignment $\uparrow$ & 0.706        & 0.597    & 0.621     & 0.623  & 0.680        & 0.640     & 0.575           & 0.657      \\
Human Preference $\uparrow$  & 4.2\%        & 10.1\%    & 2.6\%     & 5.7\%  &  6.3\%        &   21.1\%      &  7.9\%     & \textbf{42.1\%} \\
\hline
\end{tabular}
}
\vspace{-0.6cm}
\label{tab:exp2}
\end{table}

\textbf{Experiment Results} 
Following StyleShot \citep{gao2024styleshot}, we report the quantitative comparison on text and image alignment with state-of-the-art text-driven style transfer methods in Table~\ref{tab:exp2}.
% As mentioned in previous studies \citep{liu2023stylecrafter, sohn2024styledrop}, text and image alignment scores are not ideal for evaluation in style transfer tasks. 
% We present the evaluation results in this paper for reference purposes. The highest human preference scores demonstrate the effectiveness of our method.
Figure~\ref{fig:5} displays our results and baselines of four distinct style images, each corresponding to the same pair of text prompts. 
As shown in Figure~\ref{fig:5}, we observe that 
InstantStyle \citep{wang2024instantstyle} and the most recent method StyleShot \citep{gao2024styleshot} retain the image style but may fail to generate the target semantic information. 
In contrast, our method can improve text fidelity for text prompts without sacrificing style enhancement, avoiding content leakages and achieving style enhancement. 
As shown in the last column of Figure~\ref{fig:5}, 
we also present ablation study results in the last column, where we retain the identified elements to be discarded while masking the other features. Consequently, there is almost no style information identical to the reference image, further confirming that our method can efficiently and accurately decouple content from style.

\vspace{-0.2cm}
\subsection{Ablation Studies} 
\vspace{-0.2cm}

\begin{wrapfigure}{r}{0.5\textwidth}
% \begin{figure}
  \centering
  \vspace{-0.3cm}
  \includegraphics[width=\linewidth]{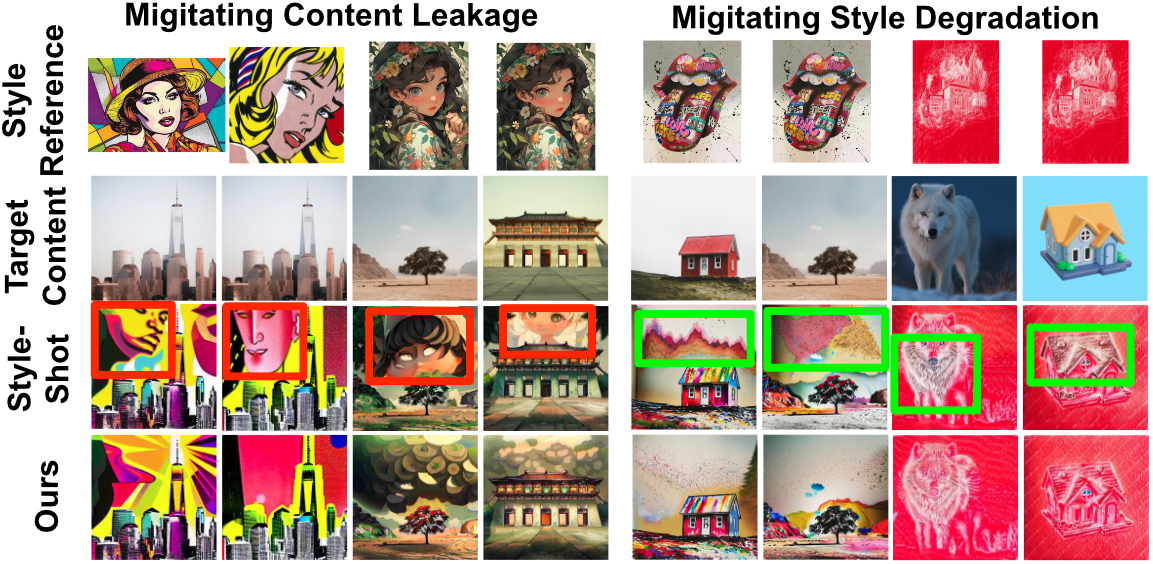}
  \vspace{-0.66cm}
   \caption{Visual comparison between StyleShot and ours in image-driven style transfer.
   Results with \textcolor{red}{content leakages} and \textcolor{green}{style degradation} are highlighted with \textcolor{red}{red} and \textcolor{green}{green} boxes, respectively. More results are in Figure~\ref{fig:19}.}
\label{fig:7}
\vspace{-0.5cm}
% \end{figure}
\end{wrapfigure}

\textbf{Effectiveness in Image-Driven Style Transfer} The proposed method also excels at transferring style onto target content images. We compare our method with the recent SOTA method StyleShot \citep{gao2024styleshot} to showcase the superiority of our method in efficiently mitigating content leakages. As shown in Figure~\ref{fig:7}, StyleShot usually generates unsatisfied results when the style reference image consists of a human face. In contrast, according to the clustering results on the element-wise product feature $\boldsymbol{e}_p$, by only masking 1-5 elements, our method can successfully mitigate content leakages and style degradation of StyleShot.

Moreover, we also compare our method with InstantStyle's block-specific injection technique based on different diffusion models and style encoders. 
Due to space limitations, we provide additional results in Appendix~\ref{app-sec: Comparison between Our Masking Strategy and InstantStyle}.

\textcolor{black}{\textbf{Ablation Studies on Clustering Number}
% In Table~\ref{tab:ablation-on-k-main}, we ablate on cluster number $K$ in the text-driven style transfer on the StyleBench dataset. Due to space limitation, we provide visualization results in Appendix~\ref{app-sec: more-results-ablation-k}. It is shown that a smaller $K$, such as $K=2$, can lead to a slightly higher text alignment score since more content-related elements in the style reference are masked. Especially in the 3D model, Anime, and Baroque art style that contains more human-related images, smaller $K$ can lead to higher text alignment scores and more efficiently avoiding content leakage.
In Table~\ref{tab:ablation-on-k-main}, we conduct ablation studies clustering number $K$ in text-driven style transfer using the StyleBench dataset. Due to space limitations, we provide additional visualization results in Appendix~\ref{app-sec: more-results-ablation-k}. The results show that a smaller $K$, such as $K=2$, can lead to higher text alignment scores, as more content-related elements in the style reference are masked. This is particularly evident in styles such as 3D models, Anime, and Baroque art, which contain more human-related images, resulting in more effective content leakage avoidance.
% In these cases, a smaller $K$ results in higher text alignment scores and more efficient avoidance of content leakage.
}

\begin{wraptable}{l}{0.4\textwidth}
\centering
\vspace{-0.6cm}
\caption{Ablation study results on clustering number $K$.}
\resizebox*{1\linewidth}{!}{
\begin{tabular}{lllll} 
\hline
                                 & \multirow{2}{*}{{$K$}}       & \multirow{2}{*}{3D Model} & \multirow{2}{*}{Anime} & \multirow{2}{*}{Baroque}  \\
                                 &                                                          &                           &                            &                           \\ 
\hline
\multirow{4}{*}{\begin{tabular}[c]{@{}l@{}}image \\alignment\end{tabular}} & 2                                                      & 0.474                     & 0.372                      & 0.384                     \\
                                 & 3                                                      & 0.478                     & 0.381                      & 0.393                     \\
                                 & 4                                                      & 0.485                     & 0.390                      & 0.404                     \\
                                 & 5                                                      & 0.487                     & 0.380                      & 0.411                     \\
\hline
\multirow{4}{*}{\begin{tabular}[c]{@{}l@{}}text \\alignment\end{tabular}}  & 2                                                      & 0.213                     & 0.234                      & 0.257                     \\
                                 & 3                                                      & 0.206                     & 0.232                      & 0.253                     \\
                                 & 4                                                      & 0.189                     & 0.231                      & 0.253                     \\
                                 & 5                                                      & 0.188                     & 0.229                      & 0.252                     \\
\hline
\end{tabular}
}
\label{tab:ablation-on-k-main}
\vspace{-0.3cm}
\end{wraptable}

\vspace{-0.2cm}
\section{Conclusion}
\vspace{-0.2cm}
% Building upon the advanced IP-Adapter framework,
% we initially present preliminary experiments and observations that shed light on the underlying reasons for the challenges of style transfer. 
% Our findings indicate that models conditioned on fewer conditions outperform those conditioned on a larger, yet less coherent set of attributes. Motivated by these observations, we propose zero-shot and fine-tuning algorithms to capture good-quality style features without introducing image-text misalignment, thus achieving better text-to-image generation.
% The simulation results show that: 1) optimization on less but more consistent conditions leads to higher text fidelity and style similarity; 2) the proposed zero-shot method (i.e., masking-based strategy on image feature of style reference) can significantly improve text fidelity and reduce content leakages.
% Extensive experiment results demonstrate that our method can efficiently and effectively mitigate content leakages and style degradation.

In this paper, we propose a masking-based method, that efficiently decouples content from style without requiring tuning any model parameters. 
By masking (zeroing out) certain elements in the image feature corresponding to that content, 
we effectively eliminate content leakages from style references across various evaluation datasets.
More importantly, we have theoretically proved 
that our model, under the guidance of appropriately selected fewer conditions, achieves a smaller divergence between the generated image distribution and the real image distribution, outperforming those conditioned on larger, yet less coherent conditions. Extensive experiments across various styles and targets have demonstrated the effectiveness of our proposed method.

\section{Ethics Statement}
This work aims to make a positive impact on the field of AI-driven image generation. We aim to facilitate the creation of images with diverse styles, and we expect all related processes to comply with local laws and be used responsibly. Users should utilize this material in a way that avoids any potential bias related to sensitive attributes such as gender, race, age, and other demographic factors. 
% We believe that the responsible use of AI-driven image generation tools is essential to fostering ethical and equitable outcomes in the field.
Further details regarding our ethics statement are provided in Appendix~\ref{app-sec:ethics}

\section*{Acknowledgements}
This work is supported by National Natural Science Foundation of China under Grant (No.62106139).

\bibliography{iclr2025_conference}
\bibliographystyle{iclr2025_conference}

\clearpage
\appendix
\section{Appendix}

\subsection{Related Works}
\textbf{Stylized Image Generation}
Stylized image generation, commonly referred to as image style transfer, involves transferring the stylistic or aesthetic attributes from a reference image to a target image. 
Thanks to the significant advancements in diffusion models \citep{DDPM, SDXL, DDIM, li2024blip, rombach2022high, Classifier-Free, ramesh2022hierarchical, saharia2022photorealistic, nichol2021glide},
numerous methods \citep{sun2023sgdiff, xu2024freetuner, lu2023specialist, lin2024ctrl} have been developed to ensure style consistency across images generated. Among inversion-based approaches \citep{InST, gal2022imageworthwordpersonalizing, StyleAligned} project style images into a learnable embedding in the text token space to guide style-specific generation. Unfortunately, these methods can lead to information loss due to the mapping from visual to text modalities.
Cross-attention manipulation \citep{le2022styleid, StyleAligned, chung2024style, StyleAligned, StyleAdapter} is another method for achieving style transfer, involving the manipulation of features within self-attention layers. 
In contrast, IP-Adapter \citep{ye2023ip} and Style-Adapter \citep{StyleAdapter} introduce a distinct cross-attention mechanism that de-couples the attention layers for text and image features, allowing for a coarse control over the style transfer process. 
Although these methods have achieved significant advancements, they often struggle with content leakages from style-reference images.

\textbf{Methods Addressing Content Leakages}
Some approaches \citep{zhang2018separating, zhang2018unifiedframeworkgeneralizablestyle, qi2024deadiff} aim to tackle the content leakages issue by constructing paired datasets where images share the same subject matter but exhibit distinct styles, facilitating the extraction of disentangled style and content representations. DEADiff \citep{qi2024deadiff} stands out by extracting disentangled representations of content and style using a paired dataset, facilitated by the Q-Former \citep{li2023blip} technique.
Other works \citep{sohn2024styledrop, liu2023stylecrafter} optimize some or all model parameters using extensive style images, embedding their visual style into the model’s output domain. However, the inherently underdetermined nature of style makes the creation of large-scale paired datasets or style datasets both resource-intensive and limited in the diversity of styles it can capture. To address this issue, InstantStyle \citep{wang2024instantstyle}, a recent innovation, employs block-specific injection and feature subtraction techniques to implicitly achieve decoupling of content and style, offering a nuanced approach to style transfer. In the context of image-driven style transfer, InstantStyle-Plus \citep{wang2024instantstyle-plus} further proposed several techniques to prioritize the integrity of the content image while seamlessly integrating the target style. 
Although the InstantStyle approach achieved significant advancements, feature manipulation across different modalities inevitably introduces the image-text misalignment issue \citep{Misalign, gordon2023mismatch}, which hinders accurate disentanglement of content and style. 
StyleDiffusion \citep{wang2023stylediffusion} introduced a CLIP-based style disentanglement loss coordinated with a style reconstruction to decouple content from style in the CLIP image space. However, this framework required a training process to disentangle style from each style image, achieving this by providing approximately 50 content images for training.
DiffuseIT \citep{kwon2022diffusion} introduced a novel diffusion-based unsupervised image translation method for decoupling content from style, but it also requires complex loss regularization.
\textcolor{black}{More recent and stronger models, such as RB-Modulation \citep{rout2024rb}, have been proposed to alleviate the content leakage problem. RB-Modulation uses attention-based feature aggregation and different descriptors to decouple content and style. It is training-free and is reported to outperform InstantStyle. CSGO \citep{xing2024csgo} is another recent approach that employs a separately trained style projection layer to mitigate content leakage.}
Additionally, Zhao et al. \citep{zhao2024identifying} proposed a method to identify and address the issue of conditional content leakage in image-to-video (I2V) generation. Several studies \citep{motamed2023lego, huang2024learning, le2022styleid} focus on concept disentanglement, but they are not specifically aimed at style transfer.

\textcolor{black}{\textbf{Masking Mechanism in Synthesizing High-Quality Images}
Although several studies \citep{couairon2022diffedit, gao2023masked, hansen2024unified, pan2023masked, lei2023masked} have explored the effectiveness of masking mechanisms, our method differs from these approaches in several key aspects:
1) No coupled denoising processes: Our method avoids the need for two denoising processes, thus saving computational resources. For instance, the DIFFEDIT method \citep{couairon2022diffedit} requires two denoising processes—one conditioned on the query text and the other conditioned on a reference text. By contrasting the predictions of the two diffusion models, DIFFEDIT generates a mask that locates the regions needing editing to match the query text.
2) Masking in the latent space: Unlike DIFFEDIT \citep{couairon2022diffedit}, which operates on the pixel level to generate a mask highlighting the regions of the input image that need editing, our method performs masking in the latent space, bypassing pixel-level operations and patch-level manipulations.
3) Focus on content leakage in style transfer: While the MDT method \citep{gao2023masked} introduces a latent masking scheme to enhance the DPMs' ability to learn contextual relations among object semantics in an image, it focuses on predicting randomly masked tokens from unmasked ones. In contrast, our method targets content leakage in style transfer. We mask feature elements that are related to unwanted content from the style reference, guided by clustering results on the element-wise product. 
% The ``From Test to Mask'' method leverages the rich multi-modal knowledge embedded in diffusion models to perform segmentation. By comparing different correlation maps in the denoising U-Net, it generates the final segmentation mask.
}

\subsection{Proof}
\label{app-sec:proof}

\textbf{Proof of Proposition~\ref{proposition}}
We denote the masked elements in the image feature as $\boldsymbol{e}_1^{s+1}, \cdots, \boldsymbol{e}_1^{d}$ and denote the feature composed by these elements as $\boldsymbol{e}_1^m$, i.e., $\boldsymbol{e}_1^m :=[\boldsymbol{e}_1^{s+1}, \cdots, \boldsymbol{e}_1^{d}]$.
Following \citep{yu2023freedom}, we use a time-independent distance measuring functions $\mathcal{D}_{\boldsymbol{\theta}}(c,\mathbf{x}_0)$  to approximate the energy function $\mathcal{E}(c,\mathbf{x}_0)$:
\begin{equation}
    \mathcal{E}(c,\mathbf{x}_t)\approx\mathbb{E}_{p(\mathbf{x}_0|\mathbf{x}_t)}\mathcal{D}_{\boldsymbol{\theta}}(c,\mathbf{x}_0)
\end{equation}
where $\boldsymbol{\theta}$ defines the model parameters of the diffusion model. $\mathcal{D}_{\boldsymbol{\theta}}(c,\mathbf{x}_0)$ computes the cosine similarity between the CLIP embeddings of the given condition $c$ and image $\mathbf{x}_0$.

Based on the classifier-free guidance \citep{Classifier-Free}, 
incorporating the masking strategy, i.e., $\boldsymbol{e}_1^m=\varnothing$ leads to $\nabla_{\mathbf{x}_t}\log p(\boldsymbol{e}_1^m|\mathbf{x}_t,\boldsymbol{e}_3)=0$. 

Building upon the energy-based assumption in \citep{yu2023freedom}, i.e., 
$\nabla_{\mathbf{x}_t}\log p(c|\mathbf{x}_t)\propto-\nabla_{\mathbf{x}_t}\mathcal{E}(c,\mathbf{x}_t)$, 
we have approximated the local maximization of $\mathcal{E}(\boldsymbol{e}_1^{m},\mathbf{x}_t)$. 
Again since
$\mathcal{E}(\boldsymbol{e}_1^{m},\mathbf{x}_t)
    \approx \mathcal{D}_{\boldsymbol{\theta}}(\boldsymbol{e}_1^{m},\mathbf{x}_{0|t})$,
we have approximated the local maximization of $\mathcal{D}_{\boldsymbol{\theta}}(\boldsymbol{e}_1^{m},\mathbf{x}_{0|t})$.

Based on the clustering result on element-wise product feature $\boldsymbol{e}_p (\boldsymbol{e}_p^i = \boldsymbol{e}_1^i \cdot \boldsymbol{e}_2^i)$, we mask (drop out) the high-value elements, denoted as $\boldsymbol{e}_1^m$, in feature $\boldsymbol{e}_p$.
With a fixed masking proportion, our proposed strategy differs from other methods—such as those relying on the absolute difference between $\boldsymbol{e}_1^i$ and $ \boldsymbol{e}_2^i$—by ensuring that the selected component $[0, \cdots, 0,\boldsymbol{e}_1^{s+1}, \cdots, \boldsymbol{e}_1^{d}]$ have the highest cosine similarity with the content text feature $\boldsymbol{e}_2$. 
% Given the fixed masking proportion, compared to other masked element selection methods, like relying on the absolute difference between $\boldsymbol{e}_1^i$ and $ \boldsymbol{e}_2^i$ or other metrics,
% the proposed masking strategy enforces the selected component $[0, \cdots, 0,\boldsymbol{e}_1^{s+1}, \cdots, \boldsymbol{e}_1^{d}]$ to have the smallest cosine similarity with the content text feature $\boldsymbol{e}_2$.
This approach can lead to $\max \mathcal{D}_{\boldsymbol{\theta}}(\boldsymbol{e}_2,\mathbf{x}_{0|t})$ when compared to other masking methods.

Note that the energy function for $\boldsymbol{e}_2$ satisfies:
$\mathcal{E}(\boldsymbol{e}_2,\mathbf{x}_t)
=\mathcal{D}_{\boldsymbol{\theta}}(\boldsymbol{e}_2,\mathbf{x}_{0|t})
$.
Therefore, according to the relation between energy and distance as defined in Equation~\ref{eq:energy}, the masking strategy with clustering on $\boldsymbol{e}_1^i \cdot \boldsymbol{e}_2^i$ can lead to the highest energy score for the content text feature $\boldsymbol{e}_2$.

\textbf{Proof of Theorem~\ref{theorem:1}}
We denote the element in the image feature and text feature as $\boldsymbol{e}_1^{i} (i\in\{1, \cdots, d\})$ and $\boldsymbol{e}_2^{i} (i\in\{1, \cdots, d\})$, respectively.
Thus, the divergence between the generated and ground-truth image distribution of the InstantStyle model is:
$$
D_1 = \mathbb{E}_{q(\boldsymbol{e}_1^{i, \cdots, d}, \boldsymbol{e}_2^{i, \cdots, d}, \boldsymbol{e}_3^{i, \cdots, d})}\mathcal{D}(q(\boldsymbol{x}|\boldsymbol{e}_1, \boldsymbol{e}_2, \boldsymbol{e}_3)\|p_{\boldsymbol{\theta}}(\boldsymbol{x}|\boldsymbol{e}_1^{i, \cdots, d}, \boldsymbol{e}_2^{i, \cdots, d}, \boldsymbol{e}_3^{i, \cdots, d}))
$$
We denote the masked element in the image feature as $\boldsymbol{e}_1^{s+1}, \cdots, \boldsymbol{e}_1^d$, 
and the divergence result of the proposed masking strategy is:
$$
D_2 = \mathbb{E}_{q(\boldsymbol{e}_1^{i, \cdots, d}, \boldsymbol{e}_2^{i, \cdots, d}, \boldsymbol{e}_3^{i, \cdots, d})}\mathcal{D}(q(\boldsymbol{x}|\boldsymbol{e}_1, \boldsymbol{e}_2, \boldsymbol{e}_3)\|p_{\boldsymbol{\theta}}(\boldsymbol{x}|\boldsymbol{e}_1^{i, \cdots, s}, \boldsymbol{e}_2^{i, \cdots, d}, \boldsymbol{e}_3^{i, \cdots, d}))
$$
With the assumption:
$$\mathbb{E}_{q(\boldsymbol{e}_1^{i, \cdots, d})} \mathcal{D}(q(\boldsymbol{x}|\boldsymbol{e}_1^{i, \cdots, d})\| p_{\theta}(\boldsymbol{x}|\boldsymbol{e}_1^{i, \cdots, s})) \leq  \mathbb{E}_{q(\boldsymbol{e}_1^{i, \cdots, s})} \mathcal{D}(q(\boldsymbol{x}|\boldsymbol{e}_1^{i, \cdots, s})\| p_{\theta}(\boldsymbol{x}|\boldsymbol{e}_1^{i, \cdots, s}))$$ 
and by Jensen's inequality \citep{ash2000probability}, we have 
\begin{equation}
\begin{aligned}
    & \mathbb{E}_{q(\boldsymbol{e}_1^{i, \cdots, d}, \boldsymbol{e}_2^{i, \cdots, d}, \boldsymbol{e}_3^{i, \cdots, d})}\mathcal{D}(q(\boldsymbol{x}|\boldsymbol{e}_1, \boldsymbol{e}_2, \boldsymbol{e}_3)\|p_{\boldsymbol{\theta}}(\boldsymbol{x}|\boldsymbol{e}_1^{i, \cdots, s}, \boldsymbol{e}_2^{i, \cdots, d}, \boldsymbol{e}_3^{i, \cdots, d})) \\
    & \leq \mathbb{E}_{q(\boldsymbol{e}_1^{i, \cdots, s}, \boldsymbol{e}_2^{i, \cdots, d}, \boldsymbol{e}_3^{i, \cdots, d})}
    \mathcal{D} (\mathbb{E}_{q(\boldsymbol{e}_1^{s+1, \cdots, d})}q(\boldsymbol{x}|\boldsymbol{e}_1, \boldsymbol{e}_2, \boldsymbol{e}_3)\|
    \mathbb{E}_{q(\boldsymbol{e}_1^{s+1, \cdots, d})}p_{\boldsymbol{\theta}}(\boldsymbol{x}|\boldsymbol{e}_1^{i, \cdots, d}, \boldsymbol{e}_2^{i, \cdots, d}, \boldsymbol{e}_3^{i, \cdots, d})) \\
    & \leq \mathbb{E}_{q(\boldsymbol{e}_1^{s+1, \cdots, d})} \mathbb{E}_{q(\boldsymbol{e}_1^{i, \cdots, s}, \boldsymbol{e}_2^{i, \cdots, d}, \boldsymbol{e}_3^{i, \cdots, d})}\mathcal{D}(q(\boldsymbol{x}|\boldsymbol{e}_1, \boldsymbol{e}_2, \boldsymbol{e}_3)\|p_{\boldsymbol{\theta}}(\boldsymbol{x}|\boldsymbol{e}_1^{i, \cdots, d}, \boldsymbol{e}_2^{i, \cdots, d}, \boldsymbol{e}_3^{i, \cdots, d}))\\
    & =\mathbb{E}_{q(\boldsymbol{e}_1^{i, \cdots, d}, \boldsymbol{e}_2^{i, \cdots, d}, \boldsymbol{e}_3^{i, \cdots, d})}\mathcal{D}(q(\boldsymbol{x}|\boldsymbol{e}_1, \boldsymbol{e}_2, \boldsymbol{e}_3)\|p_{\boldsymbol{\theta}}(\boldsymbol{x}|\boldsymbol{e}_1^{i, \cdots, d}, \boldsymbol{e}_2^{i, \cdots, d}, \boldsymbol{e}_3^{i, \cdots, d}))
\end{aligned}
\end{equation}
The last line is because of the assumption that the elements of the image feature are independent of each other. 
Thus we complete the proof of $D_2 \leq D_1$.

\textbf{Proof of Theorem~\ref{theorem:2}}
Suppose the divergence $\mathcal{D}$ is convex, and the function space $\Phi$ and $\Psi$ ($\boldsymbol{\phi} \in \Phi$ and $\boldsymbol{\psi} \in \Psi$) includes all measurable functions. 
Under Assumption~\ref{assumption} and by Jensen's inequality, we have:
\begin{equation}
\begin{aligned}
    \mathcal{D}(q(\boldsymbol{x}|c_1, c_3)\|p_{\boldsymbol{\theta},\boldsymbol{\psi}}(\boldsymbol{x}|c_1, c_3)) 
& = \mathcal{D}(\mathbb{E}_{q(c_2)}q(\boldsymbol{x}|c_1, c_2,  c_3)\|\mathbb{E}_{q(c_2)}p_{\boldsymbol{\theta}, \boldsymbol{\psi}}(\boldsymbol{x}|c_1, c_2,  c_3))\\ 
& \leq \mathbb{E}_{q(c_2)}\mathcal{D}(q(\boldsymbol{x}|c_1, c_2,  c_3)\|p_{\boldsymbol{\theta},\boldsymbol{\psi}}(\boldsymbol{x}|c_1, c_2,  c_3))
\end{aligned}
 \label{eq:2}
\end{equation} 
Under Assumption~\ref{assumption}, the condition $c_1 \in \mathcal{C}_1$ and $c_3 \in \mathcal{C}_3$ are independent of each other, and the condition $c_2$ is dependent on the style reference $c_1$. 
Given the condition $c_1=x$, the content information $c_2=y$ is uniquely determined. 
Thus, we have $q(c_2)q(c_1,c_3) \geq q(c_1, c_2, c_3)$, since $q(c_1=x, c_2=y, c_3=z)=q(c_1=x,c_3=z)$ and $q(c_1=x, c_2\neq y, c_3=z)=0$.

By the Tower Law \citep{ash2000probability} and non-negativity of $\mathcal{D}$, we have 
\begin{equation}
\begin{aligned}
&\mathbb{E}_{q(c_1, c_2,  c_3)} \mathcal{D}(q(\boldsymbol{x}|c_1, c_2,  c_3)\| p_{\theta, \boldsymbol{\psi}}(\boldsymbol{x}| c_1, c_3)) \\ 
 &\leq
\mathbb{E}_{q(c_2|c_1, c_3)}\mathbb{E}_{q(c_1, c_3)} \mathcal{D}(q(\boldsymbol{x}|c_1, c_2,  c_3)\| p_{\theta, \boldsymbol{\psi}}(\boldsymbol{x}| c_1, c_3)) \\
& = \mathbb{E}_{q(c_1, c_3)}
\mathcal{D}(q(\boldsymbol{x}|c_1, c_3)\| p_{\theta, \boldsymbol{\psi}}(\boldsymbol{x} |c_1, c_3))
\end{aligned}
\end{equation}

It is straightforward to extend this to the minimization case, leading to the following inequality:
\begin{equation}
    \min_{\boldsymbol{\psi}} \mathbb{E}_{q(c_1, c_2,  c_3)}\mathcal{D}(q(\boldsymbol{x}|c_1, c_2,  c_3)\|p_{\boldsymbol{\theta},\boldsymbol{\psi}}(\boldsymbol{x}|c_1, c_3)) \leq 
    \min_{\boldsymbol{\psi}} \mathbb{E}_{q(c_1, c_3)}\mathcal{D}(q(\boldsymbol{x}|c_1, c_3)\|p_{\boldsymbol{\theta},\boldsymbol{\psi}}(\boldsymbol{x}|c_1, c_3))
    \label{eq:1}
\end{equation}
Combining Equation~\ref{eq:2} and Equation~\ref{eq:1}, we have:
\begin{equation}
\begin{aligned}
    \min_{\boldsymbol{\psi}} \mathbb{E}_{q(c_1, c_2,  c_3)}\mathcal{D}
    & (q(\boldsymbol{x}|c_1, c_2,  c_3)\|p_{\boldsymbol{\theta},\boldsymbol{\psi}}(\boldsymbol{x}|c_1, c_3)) \\
& \leq  
    \min_{\boldsymbol{\phi}} \mathbb{E}_{q(c_1, c_2,  c_3)}\mathcal{D}(q(\boldsymbol{x}|c_1, c_2,  c_3)\|p_{\boldsymbol{\theta},\boldsymbol{\phi}}(\boldsymbol{x}|c_1, c_2,  c_3)) 
\end{aligned}
\end{equation}
Thus complete the proof.

\begin{figure}[H]
  \centering
  \includegraphics[width=1.0\linewidth]{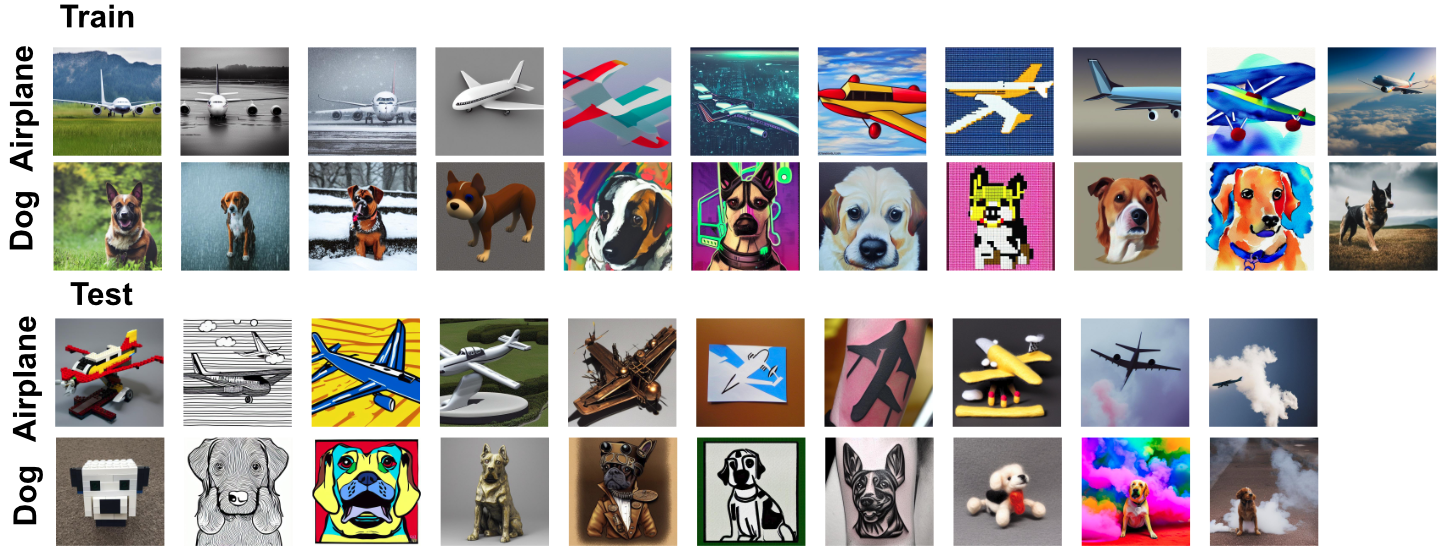}
  \vspace{-0.5cm}
   \caption{Visual examples of the style-reference images in our constructed dataset. The training sets are only used for learning Image-Adapter and Text-Adapter.}
   \label{fig:data-examples}
\end{figure}

\subsection{Algorithm}
\label{app-sec:algorithm}

\textcolor{black}{As shown in the 9th line of Algorithm~\ref{algoritm2}, we highlight the optimization objective for the Text-Adapter and Image-Adapter, respectively. These optimization objectives are used to minimize the prediction error of noise in reconstructing style reference while maximizing the difference between the model conditioned on the style reference's content and that conditioned on the target prompt.}

% \begin{minipage}[c]{.46\linewidth}  % mini-page
\begin{algorithm}[H]
   \caption{Algorithm of the proposed Tuning-Free masking-based method}
\begin{algorithmic}[1]
   \STATE {\bfseries Input:} 
   the style reference $c_1$, default style reference's content text prompt $c_2$, target text prompt $c_3$, VAE-encoder $\mathcal{E}$,
   pre-defined parameters $\alpha_t$, the repeat times of time travel $T$.
   % \STATE $(c_1,c_2, c_3)\sim p(c_1,c_2,c_3)\text{\quad}\triangleright\text{Sample conditions from data distribution}$
    \FOR {$t=1$ {\bfseries to} $T$}
    \STATE $\boldsymbol{\epsilon}\sim\mathcal{N}(\mathbf{0},\mathbf{I})$, $\mathbf{z}_0=\mathcal{E}(c_1)$
    \STATE $\mathbf{z}_t=\alpha_t\mathbf{z_{t-1}}+(1-\alpha_t)\boldsymbol{\epsilon}\text{\quad}\triangleright\text{Add noise to the latent feature}$
    \ENDFOR
    \FOR {$t=T$ {\bfseries to} 1}
    \STATE Calculate noise prediction (InstantStyle performs (a) and ours performs (b)): \\
    (a) $\boldsymbol{\epsilon}_{\theta}(\mathbf{z}_t, t,\boldsymbol{e}_1 - \boldsymbol{e}_2, \boldsymbol{e}_3)$ $\textcolor{blue}{~~~~~~\text{(InstantStyle)}}$\\
    (b) Perform clustering (such as K-means clustering \citep{Jin2010}) on element product $\boldsymbol{e}_1^i \cdot \boldsymbol{e}_2^i$; Generate masking vector $\mathbf{m}$, where we set the element value to 1 for those elements in the highest-means cluster; Calculate $\boldsymbol{\epsilon}_{\theta}(\mathbf{z}_t,t, \boldsymbol{e}_1\odot\mathbf{m}, \boldsymbol{e}_3)$ $\textcolor{green}{~~~~~~\text{(Ours)}}$\\
    \STATE
    Denoise diffusion model using predicted noise;
    % $\mathbf{z}_{t-1}=\frac1{\sqrt{\alpha_t}}\left(\mathbf{z}_t-\frac{1-\alpha_t}{\sqrt{1-\bar{\alpha}_t}}\boldsymbol{\epsilon}_{\theta}
    % % (\mathbf{z}_t,c_1\odot\mathbf{m}, c_3)
    % \right) +\sigma_t\mathbf{z}$, where $\mathbf{z}\sim\mathcal{N}(\mathbf{0},\mathbf{I})$
    \ENDFOR
   \STATE {\bfseries Output:} decode the reversed latent code $\mathbf{z}_{0}$ to image space and output the generated image
\end{algorithmic}
\label{algorithm1}
\end{algorithm}
% \end{minipage}
% \hspace{0.3cm}
%%%%%%%%%%%%%%%%%%%%%%%%%%%%%%%%%%%%%%%%%%%%%%%%%%%
% \begin{minipage}[c]{.49\linewidth}  % mini-page
\begin{algorithm}[H]
\caption{Algorithm of the Tuning-Based model: Image-Adapter and Text-Adapter}
\begin{algorithmic}[1]
   \STATE {\bfseries Input:} the data distribution of style reference $c_1$, style reference's content text prompt $c_2$ and target text prompt $c_3$, 
   text adapter $\boldsymbol{\phi}_{\theta_1}(\cdot)$, image adapter $\boldsymbol{\psi}_{\theta_2}(\cdot)$, VAE-encoder $\mathcal{E}$,
   pre-defined parameters $\alpha_t$, hyper-parameter $\lambda$,
   maximum training step $M$ and the repeat times of time travel $T$.
   \FOR{$m=1$ {\bfseries to} $M$}
   \STATE $(c_1,c_2, c_3)\sim p(c_1,c_2,c_3)\text{\quad}\triangleright\text{Sample conditions from data distribution}$
    \FOR {$t=1$ {\bfseries to} $T$}
    \STATE $\boldsymbol{\epsilon}\sim\mathcal{N}(\mathbf{0},\mathbf{I})$, $\mathbf{z}_0=\mathcal{E}(c_1)$
    \STATE $\mathbf{z}_t=\alpha_t\mathbf{z_{t-1}}+(1-\alpha_t)\boldsymbol{\epsilon}\text{\quad}\triangleright\text{Add noise to the latent feature}$
    \ENDFOR
    \FOR {$t=T$ {\bfseries to} 1}
    \STATE Take the gradient step (Text-Adapter performs (a) and Image-Adapter performs (b)): \\
    (a) $\nabla_{\theta_1}[\left\|\boldsymbol{\epsilon}_{\theta}(\mathbf{z}_t,t,\boldsymbol{e}_1 - \boldsymbol{\phi}_{\theta_1}({\boldsymbol{e}_2}), \boldsymbol{e}_3)-\boldsymbol{\epsilon}\right\|^{2} 
    - \lambda  \|\boldsymbol{\epsilon}_{\theta}(\mathbf{z}_t,t,\boldsymbol{e}_1 - \boldsymbol{\phi}_{\theta_1}({\boldsymbol{e}_2}),\boldsymbol{e}_3)
    - \boldsymbol{\epsilon}_{\theta}(\mathbf{z}_t,t,\boldsymbol{e}_1 - \boldsymbol{\phi}_{\theta_1}(\boldsymbol{e}_2), \boldsymbol{e}_2)\|^{2}]$  
    $\textcolor{blue}{~~~~~~~\text{(Text-Adapter)}}$ \\
    (b)$\nabla_{\theta_2}[\left\|\boldsymbol{\epsilon}_{\theta}(\mathbf{z}_t,t,\boldsymbol{e}_1 - \boldsymbol{\psi}_{\theta_2}({\boldsymbol{e}_1}), \boldsymbol{e}_3)-\boldsymbol{\epsilon}\right\|^{2}
    - \lambda  \|\boldsymbol{\epsilon}_{\theta}(\mathbf{z}_t,t,\boldsymbol{e}_1 - \boldsymbol{\psi}_{\theta_2}({\boldsymbol{e}_1}),\boldsymbol{e}_3)-\boldsymbol{\epsilon}_{\theta}(\mathbf{z}_t, t,\boldsymbol{e}_1 - \boldsymbol{\psi}_{\theta_2}(\boldsymbol{e}_1), \boldsymbol{e}_2)\|^{2}]$  
    $\textcolor{green}{~~~~~~~\text{(Image-Adapter)}}$ \\
    \STATE Update the model parameter $\theta_1$ or $\theta_2$
    \ENDFOR
   \ENDFOR
   \STATE {\bfseries Output:} The Image-Adapter and Text-Adapter model
\end{algorithmic}
\label{algoritm2}
\end{algorithm}
% \end{minipage}

\subsection{Evaluation Datasets and Metrics}
\label{app-sec: datasets-in-exp1}

\subsubsection{Evaluation Datasets}
Previous evaluation datasets do not contain explicitly defined references' contents, thus making it inaccurate in evaluating content leakages. Instead, we consider an evaluation dataset comprising various defined reference contents and styles for comprehensively assessing models' capability in addressing content leakages.
In this paper, we construct the evaluation dataset consisting of 10 content objects and 21 image styles. We present detailed information on the constructed dataset in Table~\ref{tab:datasets-in-exp1}.
Visual examples of the style-reference images are provided in Figure~\ref{fig:data-examples}.

\subsubsection{Evaluation Metrics}
Following previous studies \citep{sohn2024styledrop, gao2024styleshot}, we report the image alignment and text alignment scores in Table~\ref{tab:exp2} and Figure~\ref{fig:tuning-based-model-results}.

\textbf{Image alignment and text alignment $\uparrow$}:
Image alignment refers to the cosine similarity between the CLIP embeddings of the generated images and the style reference images, while text alignment measures the cosine similarity between the CLIP embeddings of the generated images and the target text prompts.

Our method is specifically designed to address content leakage in style transfer diffusion models. 
In addition to the image alignment and text alignment scores used in previous studies \citep{sohn2024styledrop, gao2024styleshot}, we also introduce three quantitative metrics to comprehensively assess the quality of the generated images from the perspectives of style similarity, text fidelity, and content leakages from style references.
We denote the CLIP image feature of the generated image and the style-reference image as $\boldsymbol{e}_g$ and $\boldsymbol{e}_1$, respectively.
Given the class name of style reference's content object and the target text prompt, we denote the CLIP text feature of reference's content object and text prompt as $\boldsymbol{e}_2$ and $\boldsymbol{e}_3$, respectively.

\textbf{Fidelity score $\uparrow$}:
We perform binary classification on the generated images to differentiate between the reference's content object and the text prompt, computing the classification accuracy, referred to as the {fidelity score $\uparrow$}. 
\textcolor{black}{Specifically, we denote the cosine similarity between the CLIP image feature of the generated image and the CLIP text feature of reference's content object as $\frac{\langle e_2, e_g \rangle}{|e_2| \cdot |e_g|}$. Similarly, we denote the cosine similarity between the CLIP image feature of the generated image and the CLIP text feature of text prompt as $\frac{\langle e_3, e_g \rangle}{|e_3| \cdot |e_g|}$. If $\frac{\langle e_2, e_g \rangle}{|e_2| \cdot |e_g|} < \frac{\langle e_3, e_g \rangle}{|e_3| \cdot |e_g|}$, the generated image is considered correctly classified, meaning it contains the target content rather than the content of the style reference.}
Therefore, the fidelity score mainly reflects the models' control ability of text prompts.

\textbf{Leakage score $\downarrow$}:
We calculate the similarity between the generated images and the reference's content text, calibrated by the similarity between the reference images and the content text, termed the {leakage score $\downarrow$}. Thus, the leakage score is calculated by:
% ${< \boldsymbol{e}_g,  \boldsymbol{e}_2>}/{< \boldsymbol{e}_3,  \boldsymbol{e}_2>}$.
\begin{equation}
    \begin{cases}{< \boldsymbol{e}_g,  \boldsymbol{e}_2>}/{< \boldsymbol{e}_1,  \boldsymbol{e}_2>}& \text{if $\boldsymbol{e}_g$ is accurately classified}\\1&\text{else}
\end{cases}
\end{equation}
The leakage score indicates content leakages from the style reference, where a lower score is preferable.

\textbf{Style score $\uparrow$}:
Finally, we assess the similarity between the generated images and the style reference, adjusting by subtracting the similarity between the generated images and the reference's content text, which we refer to as the {style score $\uparrow$}. 
The style score is calculated as follows:
% ${< \boldsymbol{e}_g, \boldsymbol{e}_1>} - {< \boldsymbol{e}_g, \boldsymbol{e}_2>}$.
\begin{equation}
    \begin{cases}{< \boldsymbol{e}_g, \boldsymbol{e}_1>} - {< \boldsymbol{e}_g, \boldsymbol{e}_2>}&\text{if $\boldsymbol{e}_g$ is accurately classified}\\0&\text{else}
\end{cases}
\end{equation}
Compared to the image alignment score, the style score more accurately reflects the style similarity between the generated images and the style references.

\textbf{Human preference score $\uparrow$}: In addition to objective evaluations, we have also designed a user study to subjectively assess the practical performance of various methods. 
\textcolor{black}{In Section 4.1, the constructed dataset consists of 10 content objects (from CIFAR-10) and 21 image styles (11 for training and 10 for testing) for each content object, with 8 variations per style. This results in a total of $10 \times 11 \times 8 = 880$ style references. For each style reference, we perform style transfer for 5 target text prompts, with 4 generations per target text prompt, leading to $880 \times 4 = 3520$ generations per text prompt. We randomly sample 50 images from the 3520 generated images for each target text prompt. In total, this gives us $50 \times 5 = 250$ images from each method to evaluate. The same procedure is applied in the evaluation presented in Section 4.2.}
We asked 10 users from diverse backgrounds to evaluate the generated results in terms of text fidelity, content leakages and style similarity, and to provide their overall preference considering these three aspects. 
Finally, the final average results are displayed in Table~\ref{tab:exp1-overall} and Table~\ref{tab:exp2}.

\begin{table}
\centering
\caption{Evaluation Datasets in Section~\ref{sec:exp1}.}
\vspace{0.1cm}
\resizebox*{1.0\linewidth}{!}{
\begin{tabular}{l|l|llllll} 
\hline\hline
\multirow{7}{*}{\textbf{Train}} & \multirow{6}{*}{\textbf{Style References}} & \multicolumn{5}{c}{\textbf{Content}}                                                                       \\ 
\cline{3-8}
                               &   &   An automobile                                         & An airplane                 & A bird                      & A cat      & A deer              & A dog       \\
                                &  &   A horse                                       & A frog                      & A ship                      & A truck    &                     &             \\ 
\cline{3-8}
                                &                                            & \multicolumn{5}{c}{\textbf{Styles}}                                                                        \\ 
\cline{3-8}
                                &  & natural environment      & rainy day   & snowy day                   & 3D model   & abstract            & cyberpunk   \\
                                &                                            & oil painting                & realism                     & watercolor & beautiful landscape & watercolor  \\ 
\cline{2-8}
                                & \textbf{Text Prompts}                      & \multicolumn{5}{l}{Image reconstruction: use the content of style reference}                               \\ 
\hline
\multirow{8}{*}{\textbf{Test}}  & \multirow{6}{*}{\textbf{Style References}} & \multicolumn{5}{c}{\textbf{Content}}                                                                       \\ 
\cline{3-8}
                                &  &   An automobile                                                & An airplane                 & A bird                      & A cat      & A deer              & A dog       \\
                                &  &   A horse                                             & A frog                      & A ship                      & A truck    &                     &             \\ 
\cline{3-8}
                                &                                            & \multicolumn{5}{c}{\textbf{Styles}}                                                                        \\ 
\cline{3-8}
                                &                                            & lego toy                    & statue                      & steampunk  & stick figure        & tattoo  & line art      \\
                                &                                      & wool toy                    & surround with colored smoke & pop art    &  surround with white smoke                   &              \\ 
\cline{2-8}
                                & \textbf{Text Prompts}                      & \multicolumn{1}{c}{A bench} & A human                     & A laptop   & A rocket            &  A flower           \\
\hline\hline
\end{tabular}
}
\label{tab:datasets-in-exp1}
\vspace{-0.4cm}
\end{table}

\subsection{Coefficient-Tuning Results}
\label{app-sec: more-results-exp0}

\begin{figure}[t]
  \centering
  \includegraphics[width=1.0\linewidth]{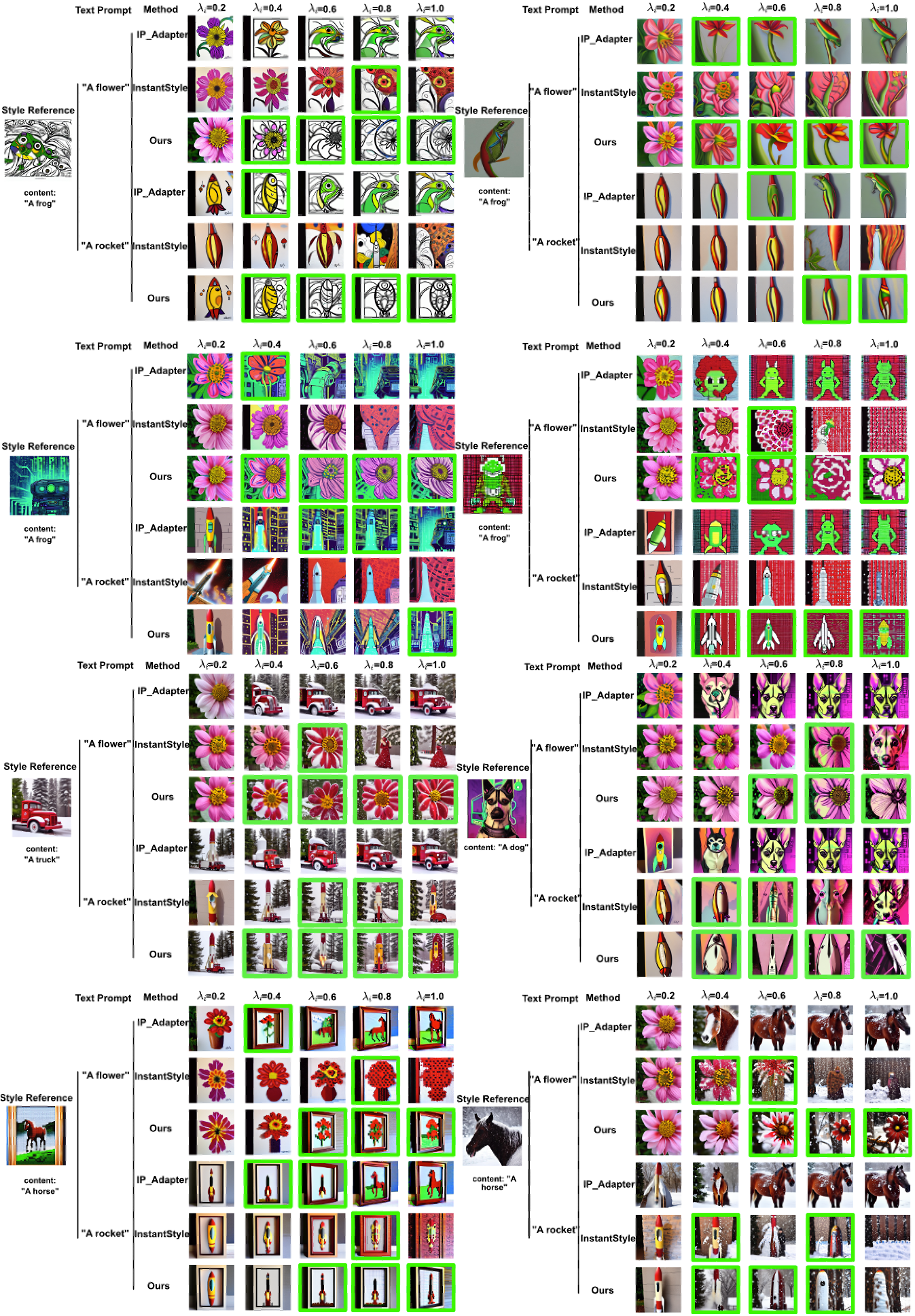}
   \caption{
   \textbf{The coefficient-tuning results of IP-Adapter \citep{ye2023ip}, InstantStyle \citep{wang2024instantstyle} and our proposed method.} We use different coefficients for the image condition (i.e., $\lambda_{i}$ in Equation~\ref{eq:linear-weighting}). We highlight the satisfactory results with \textcolor{green}{green boxes}.
   }
   \label{fig: more-results-exp0}
\end{figure}

We present the coefficient-tuning results of the IP-Adapter and InstantStyle in Figure~\ref{fig: more-results-exp0}.
For IP-Adapter and InstantStyle, lowering the coefficient for image condition helps to enhance the control ability of text prompts, but it also comes with style degradation.  
% We apply different scales to the content text feature in the feature subtraction process of InstantStyle. 
% Notably, lowering the coefficient scale helps to enhance the control ability of text prompts, but it also comes with style degradation. 
Moreover, the coefficient-tuning process is quite labour-intensive. In contrast, our method can achieve much more stable results across various coefficient values. Figure~\ref{fig: more-results-exp0} highlights the limitations of the InstantStyle approach in decoupling content and style, particularly regarding labour-intensive coefficient tuning.

\begin{figure}[!t]
  \centering
  \includegraphics[width=0.82\linewidth]{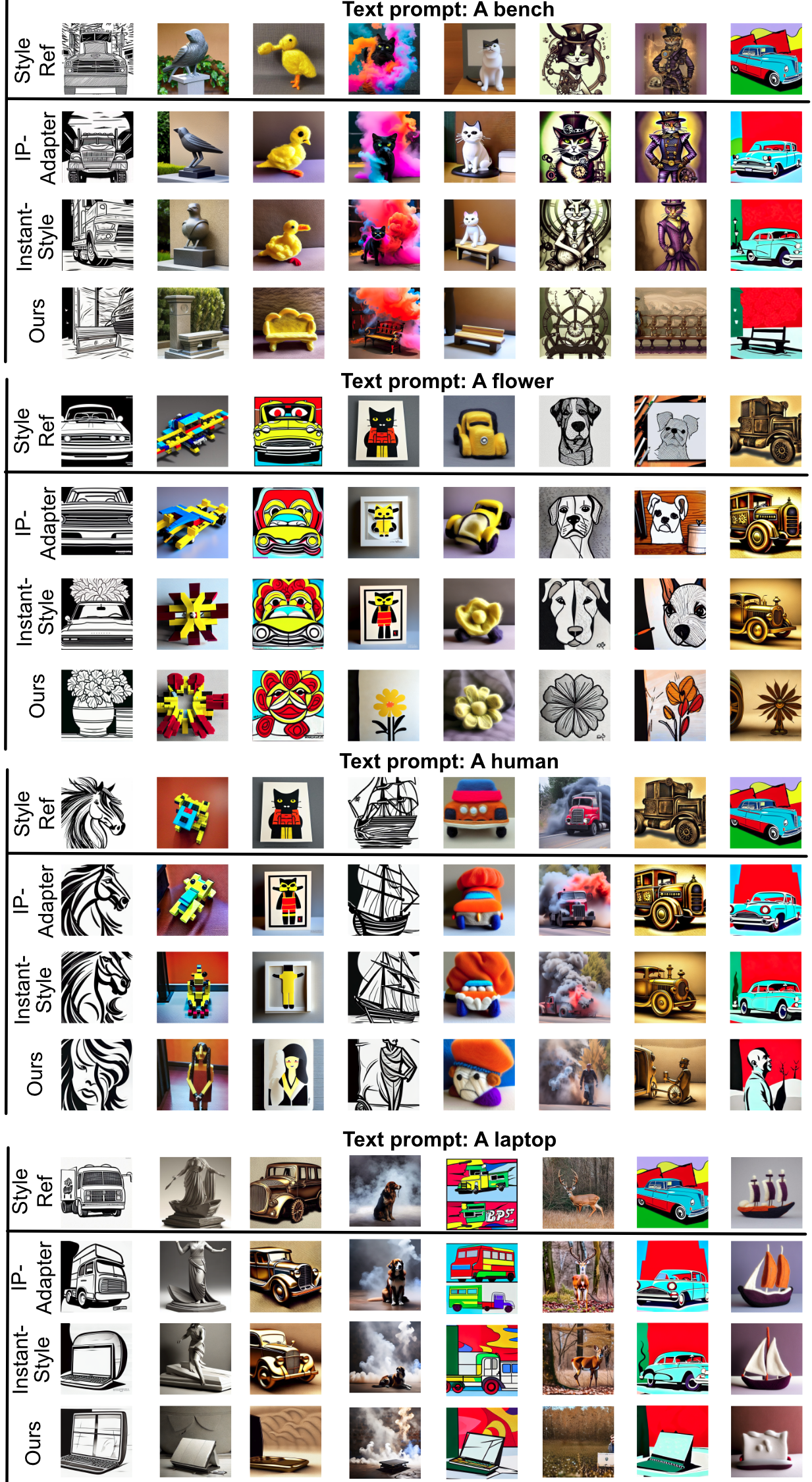}
  \vspace{-0.2cm}
   \caption{\textbf{Effectiveness of the proposed masking strategy over feature subtraction in avoiding content leakages.}
   The superior performance of our method showcases that the appropriately-selected fewer conditions can more efficiently avoid content leakages.}
   \label{fig:more-results-exp1-1}
\end{figure}

\subsection{More Visualization Results for Section~\ref{sec:experiments}}
\label{app-sec: more-results-exp1}

\begin{figure}[t]
  \centering
  \includegraphics[width=1.0\linewidth]{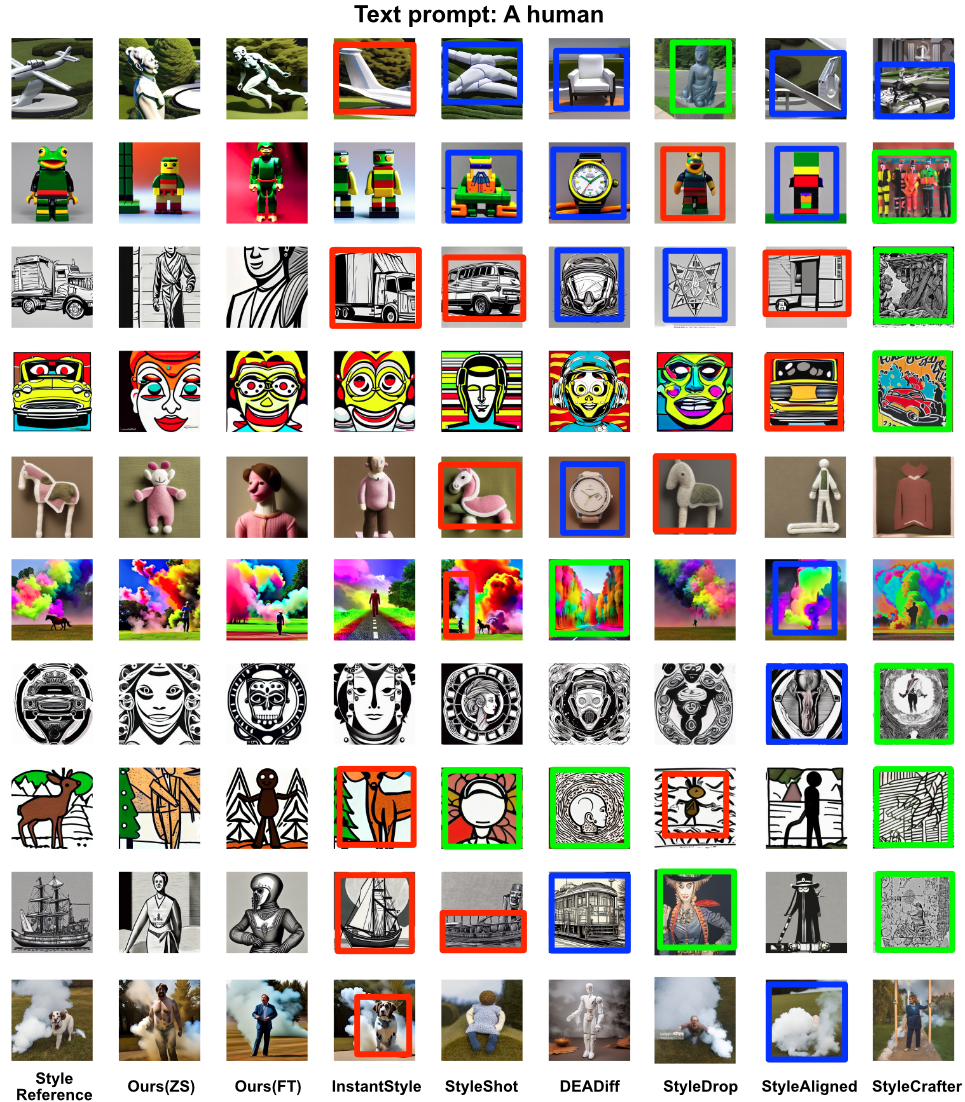}
   \caption{ Visual comparison between recent state-of-the-art methods and ours in text-driven style transfer, where the text prompt is ``A human''.}
   \label{fig:more-results-exp1-2}
\end{figure}

\begin{figure}[t]
  \centering
  \includegraphics[width=1.0\linewidth]{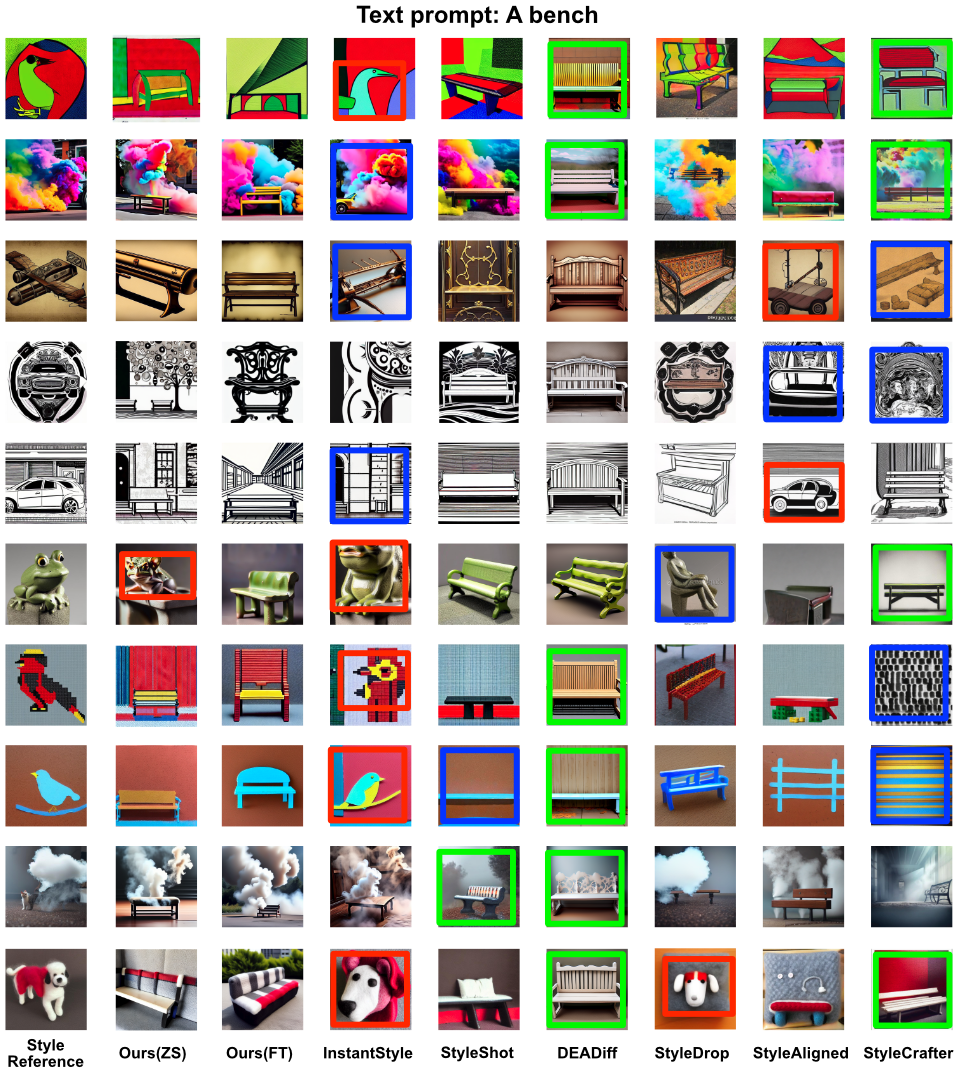}
   \caption{Visual comparison between recent state-of-the-art methods and ours in text-driven style transfer, where the text prompt is ``A bench''.}
   \label{fig:more-results-exp1-3}
\end{figure}

\begin{figure}[t]
  \centering
  \includegraphics[width=1.0\linewidth]{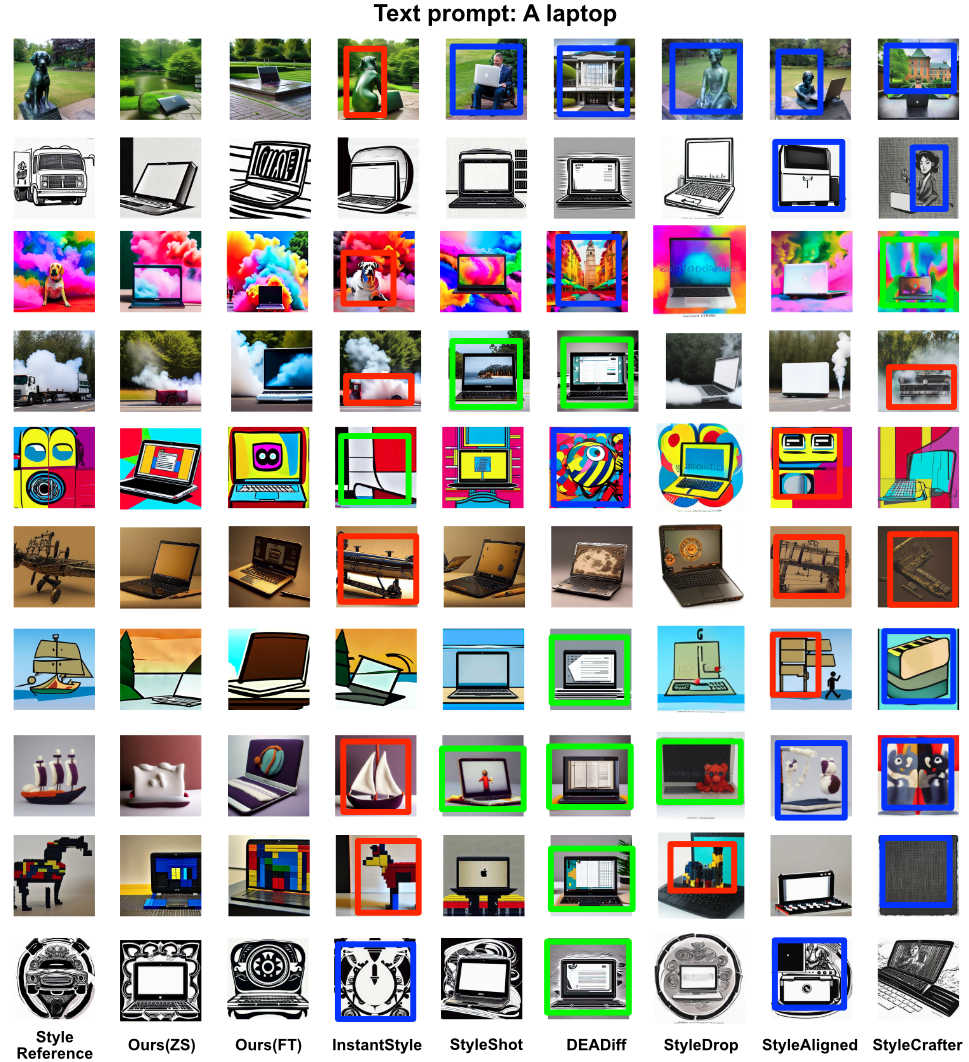}
   \caption{Visual comparison between recent state-of-the-art methods and ours in text-driven style transfer, where the text prompt is ``A laptop''.}
   \label{fig:more-results-exp1-4}
\end{figure}

\begin{figure}[t]
  \centering
  \includegraphics[width=1.0\linewidth]{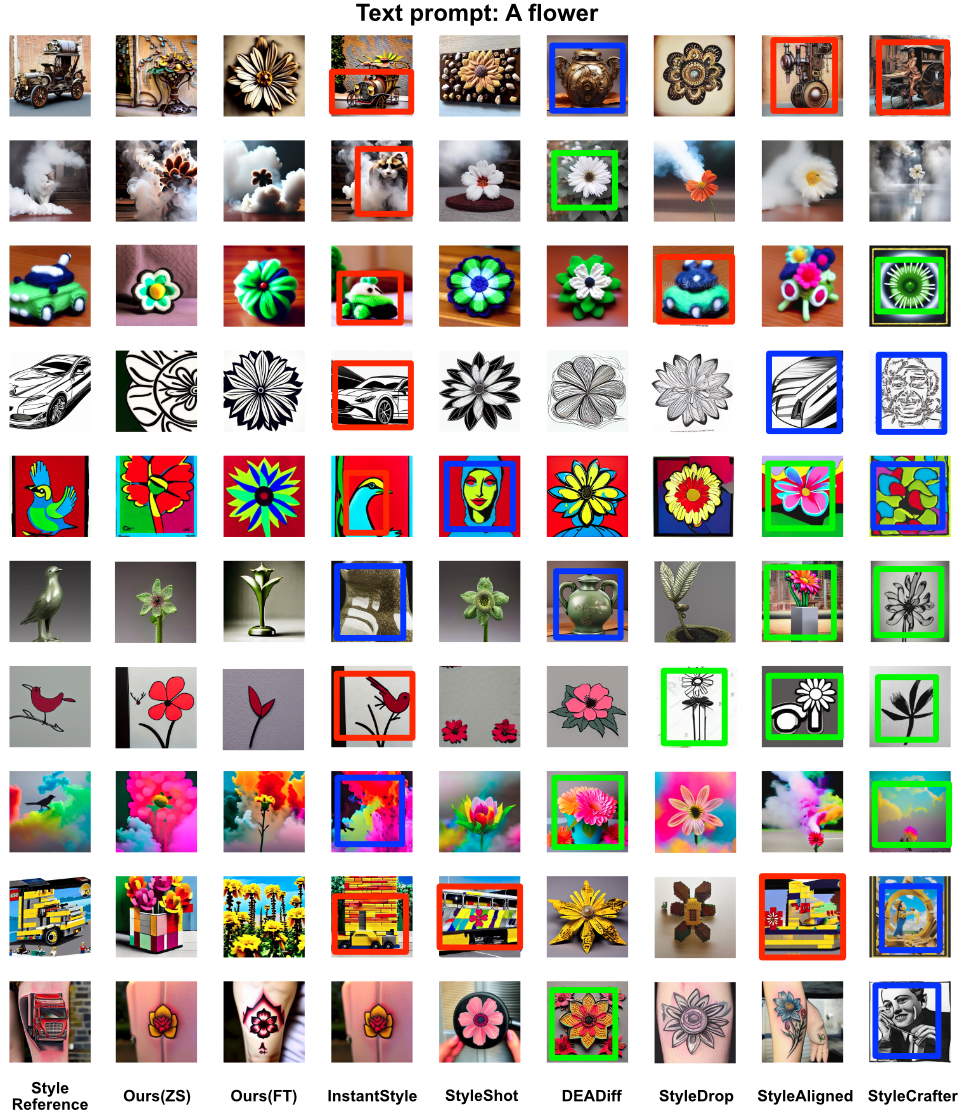}
   \caption{ Visual comparison between recent state-of-the-art methods and ours in text-driven style transfer, where the text prompt is ``A flower''.}
   \label{fig:more-results-exp1-5}
\end{figure}

\begin{figure}[t]
  \centering
  \includegraphics[width=1.0\linewidth]{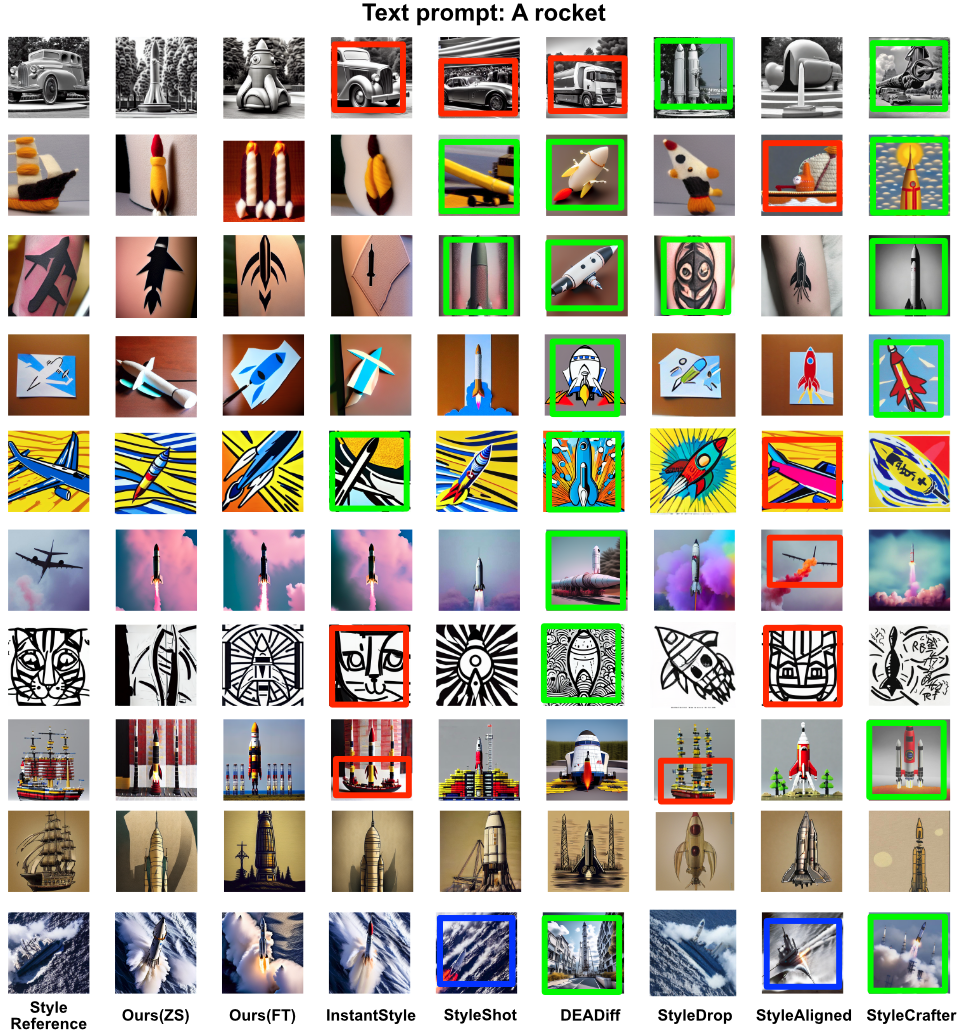}
   \caption{ Visual comparison between recent state-of-the-art methods and ours in text-driven style transfer, where the text prompt is ``A rocket''.}
   \label{fig:more-results-exp1-6}
\end{figure}

We present more visualization results for Section~\ref{sec:experiments}.
In Figure~\ref{fig:more-results-exp1-2}-\ref{fig:more-results-exp1-6}, we mark the results with significant \textcolor{red}{content leakages}, \textcolor{green}{style degradation}, and \textcolor{blue}{loss of text fidelity} with \textcolor{red}{red}, \textcolor{green}{green}, and \textcolor{blue}{blue} boxes, respectively.

\begin{figure}
  \centering
  \includegraphics[width=1\linewidth]{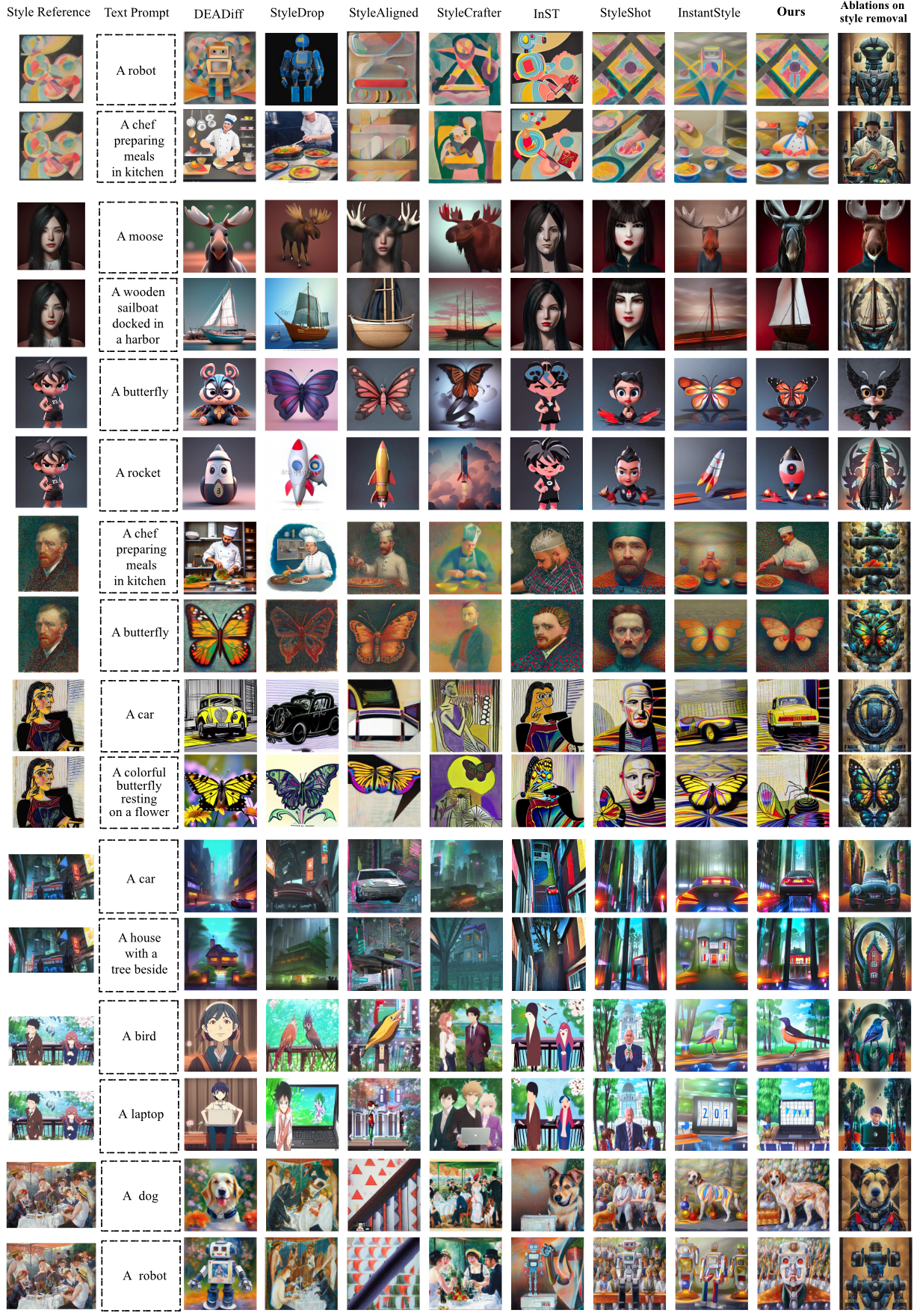}
  \vspace{-0.6cm}
   \caption{
   \textbf{Visual comparison between recent state-of-the-art methods and ours in text-driven style transfer on StyleBench.}
   The proposed masking-based method does not require content knowledge of the image reference; instead, we leverage the CLIP text feature of ``person, animal, plant, or object in the foreground'' to identify the elements that need to be masked.
   % We also present ablation study results in the last column, where we retain the identified elements to be discarded while masking the remaining features. Consequently, there is almost no style information identical to the reference image, demonstrating that our method can efficiently discard content-related elements while preserving the style-related information of style references.
   % The ablation study results in the last column demonstrate that our masking-based method can efficiently and accurately decouple content from style. 
   }
\vspace{-0.5cm}
\label{fig:more-results-on-StyleBench}
\end{figure}

% \begin{table}[H]
% \centering
% \vspace{-0.3cm}
% \caption{The quantitative analysis on content leakages between different proportions of masked elements. The model becomes the original IP-Adapter when the proportion is 0\%.}
% \vspace{0.1cm}
% \resizebox*{1.0\linewidth}{!}{
% \begin{tabular}{llllllllllll|l} 
% \hline
%    Proportion      & 0\%  & 5\%  & 10\%  & 15\%  & 20\%  & 25\%  & 30\%  & 35\%  & 40\%  & 45\%  & 50\%  & InstantStyle \\ 
% \hline
% image score $\uparrow$   & 0.605 & 0.527 & 0.508 & 0.494 & 0.483 & 0.475 & 0.469 & 0.462 & 0.459 & 0.458 & 0.454  \\
% text score $\uparrow$    & 0.204 & 0.226 & 0.231 & 0.234 & 0.236 & 0.237 & 0.237 & 0.239 & 0.238 & 0.238 & 0.238  & 0.220\\
% content score $\downarrow$ & 0.267 & 0.239 & 0.227 & 0.219 & 0.214 & 0.209 & 0.207 & 0.204 & 0.203 & 0.202 & 0.202  & 0.246\\
% \hline
% \end{tabular}
% }
% \label{tab:abaltion-on-theta}
% \vspace{-0.6cm}
% \end{table}

% \begin{figure}[t]
%   \centering
%   \includegraphics[width=1.0\linewidth]{figure/exp1_curve.png}
%    \caption{Experiment results of Theorem 1. We also compare InstantStyle with the proposed method.}
% \end{figure}

\subsection{Comparison between Our Masking Strategy and InstantStyle}
\label{app-sec: Comparison between Our Masking Strategy and InstantStyle}
%%%%%%%%%%%%%%%%%%%%%%%%%%%%%%%%%%%%%%%%%%%%%%%%%%%%%%%%%%%
\begin{figure}[!t]
  \centering
  \includegraphics[width=1.0\linewidth]{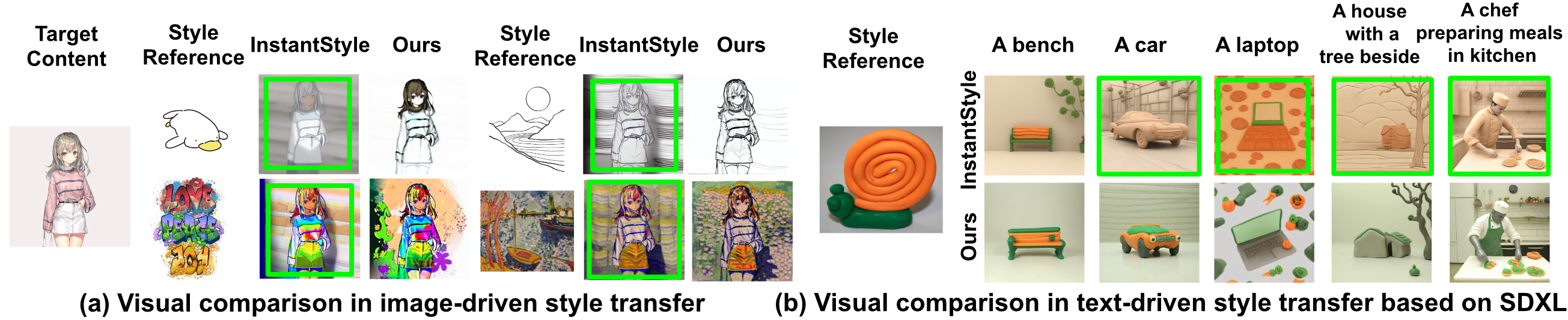}
  \vspace{-0.6cm}
   \caption{(a) Visual comparison between InstantStyle's feature subtraction and ours in image-driven style transfer. 
   % For content that needs to be removed, we use the text prompt:
   % ``person, animal, plant, or object in the foreground''.
   (b) Visual comparison between InstantStyle's block-specific injection techniques and ours in text-driven style transfer based on SDXL. 
   % We mark \textcolor{red}{content leakages} and \textcolor{green}{style degradation} of StyleShot with \textcolor{red}{red} and \textcolor{green}{green} boxes, respectively.
   }
   \label{fig:8}
\vspace{-0.6cm}
\end{figure}

To comprehensively demonstrate the effectiveness of our method in image-driven style transfer, \textcolor{black}{we compare InstantStyle's feature subtraction (i.e., directly subtracting the content text feature from the image feature) and our masking-based method based on StyleShot's style encoder,} providing visualization results in Figure~\ref{fig:8}~(a).
Due to image-text misalignment, InstantStyle may disrupt the style information extracted by StyleShot's style encoder. In contrast, guided by appropriately-selected fewer conditions and without introducing additional content text features during the denoising process, our method successfully preserves the style features.
Furthermore, we conduct ablation studies on the backbone of the diffusion model, as shown in Figure~\ref{fig:8}~(b). Using the SDXL diffusion model \citep{SDXL}, we compare our method with InstantStyle's block-specific injection technique, which injects image features into style-specific blocks in the text-driven style transfer task. While InstantStyle encounters style disruption, our method significantly alleviates this problem by leveraging fewer appropriately-selected conditions.
%%%%%%%%%%%%%%%%%%%%%%%%%%%%%%%%%%%%%%%%%%%%%%%%%%%%%%%%%%%

\subsection{Ablation studies on clustering number}
\label{app-sec: more-results-ablation-k}

\begin{figure}[t]
  \centering
  \includegraphics[width=1.0\linewidth]{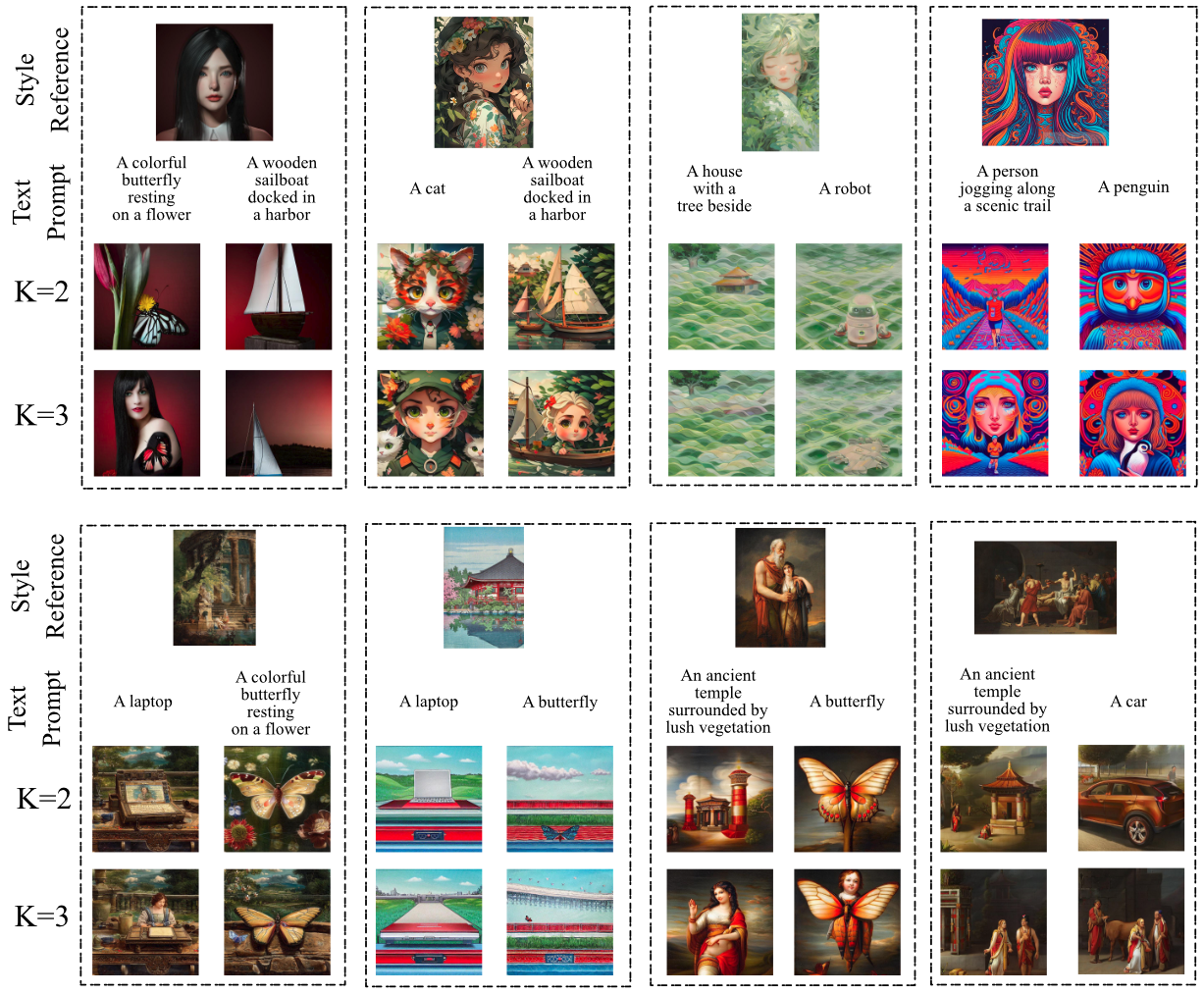}
  \vspace{-0.6cm}
   \caption{\textcolor{black}{Visual comparison of the proposed masking-based method with varying cluster numbers. It is shown that a smaller $K$, such as $K=2$, can lead to better performance in avoiding content leakage, as more content-related elements in the style reference are masked. This is particularly evident in styles such as 3D models, Anime, and Baroque art, which contain more human-related images.
In these cases, a smaller $K$ results in higher text alignment scores and more effective avoidance of content leakage.}
   }
   \label{fig:ablation_k}
\end{figure}

\textcolor{black}{
We ablate on cluster number in the text-driven style transfer based on the StyleBench dataset. We report the image alignment and text alignment results based on three different CLIP backbones in Table~\ref{table:ablation-on-k} and provide the visual comparisons of our masking-based method with varying cluster numbers in Figure~\ref{fig:ablation_k}. It is shown that a smaller $K$, such as $K=2$, can lead to better performance in avoiding content leakage, as more content-related elements in the style reference are masked. This is particularly evident in styles such as 3D models, Anime, and Baroque art, which contain more human-related images. In these cases, a smaller $K$ results in higher text alignment scores and more effective avoidance of content leakage.}

\begin{table}
\centering
\caption{Ablation study results on clustering number $K$.}
\vspace{0.1cm}
\resizebox*{0.8\linewidth}{!}{
\begin{tabular}{lllll} 
\hline
                                 & \multirow{2}{*}{\diagbox{$K$}{CLIP Model~}} & \multirow{2}{*}{ViT-B/32} & \multirow{2}{*}{ViT-L/14}  & \multirow{2}{*}{ViT-H/14}    \\
                                 &                                                          &                           &                            &                           \\ 
\hline
\multirow{4}{*}{image alignment} & 2                                                      & 0.657                     & 0.608                      & 0.403                     \\
                                 & 3                                                      & 0.656                     & 0.611                      & 0.410                     \\
                                 & 4                                                      & 0.657                     & 0.615                      & 0.415                     \\
                                 & 5                                                      & 0.657                     & 0.614                      & 0.415                     \\
\hline
\multirow{4}{*}{text alignment}  & 2                                                      & 0.265                     & 0.212                      & 0.258                     \\
                                 & 3                                                      & 0.264                     & 0.211                      & 0.253                     \\
                                 & 4                                                      & 0.265                     & 0.210                      & 0.252                     \\
                                 & 5                                                      & 0.264                     & 0.210                      & 0.252                     \\ 
\hline
                                 & \multirow{2}{*}{\diagbox{$K$}{Style}}       & \multirow{2}{*}{3D Model} & \multirow{2}{*}{Anime} & \multirow{2}{*}{Baroque}  \\
                                 &                                                          &                           &                            &                           \\ 
\hline
\multirow{4}{*}{\begin{tabular}[c]{@{}l@{}}image \\alignment \\ based on ViT-H/14\end{tabular}} & 2                                                      & 0.474                     & 0.372                      & 0.384                     \\
                                 & 3                                                      & 0.478                     & 0.381                      & 0.393                     \\
                                 & 4                                                      & 0.485                     & 0.390                      & 0.404                     \\
                                 & 5                                                      & 0.487                     & 0.380                      & 0.411                     \\
\hline
\multirow{4}{*}{\begin{tabular}[c]{@{}l@{}}text \\alignment \\ based on ViT-H/14\end{tabular}}  & 2                                                      & 0.213                     & 0.234                      & 0.257                     \\
                                 & 3                                                      & 0.206                     & 0.232                      & 0.253                     \\
                                 & 4                                                      & 0.189                     & 0.231                      & 0.253                     \\
                                 & 5                                                      & 0.188                     & 0.229                      & 0.252                     \\
\hline
\end{tabular}
}
\label{table:ablation-on-k}
\vspace{-0.4cm}
\end{table}

\begin{figure}
  \centering
  \includegraphics[width=\linewidth]{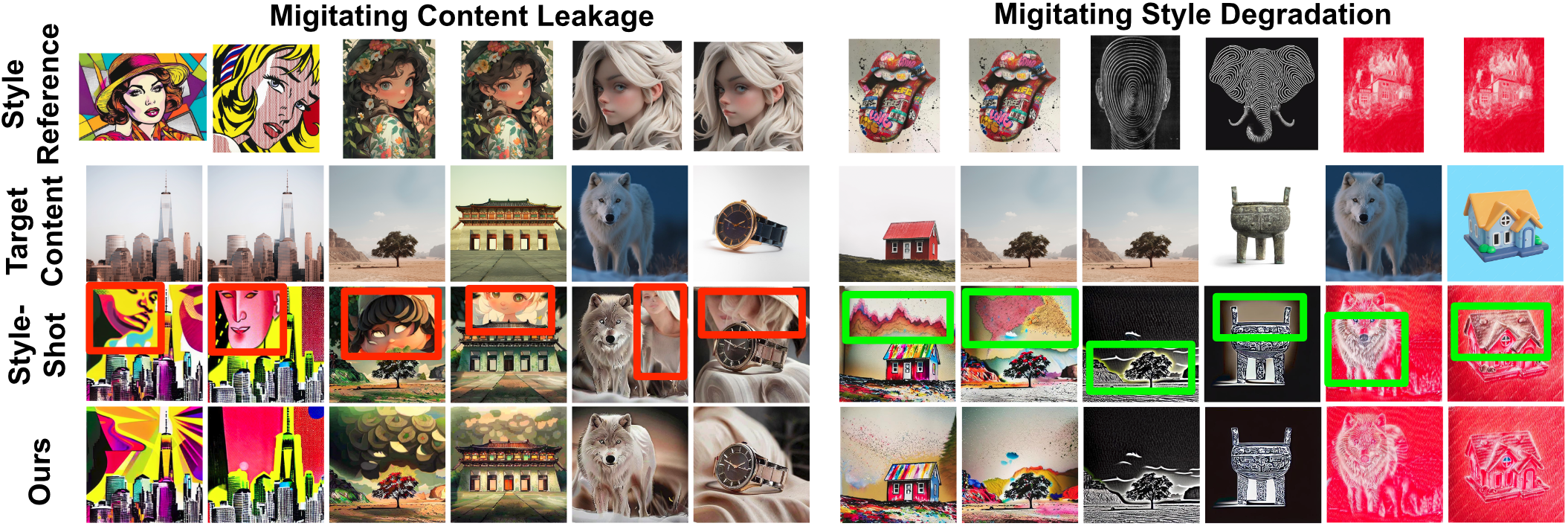}
  \vspace{-0.5cm}
   \caption{Visual comparison between StyleShot and ours in image-driven style transfer.
   Results with \textcolor{red}{content leakages} and \textcolor{green}{style degradation} are highlighted with \textcolor{red}{red} and \textcolor{green}{green} boxes, respectively. }
\label{fig:19}
\end{figure}

\subsection{Simulation Results of Proposition 1}
\label{app-sec: simulation-p1}
\textcolor{black}{
We conduct simulation experiments based on our constructed dataset to demonstrate Proposition 1. Using the energy score proposed by \citep{liu2021energy}, we calculate the energy scores of the masked image features for two different masking approaches: one based on clustering the product of $e_1^i$ and $e_2^i$ ($e_1^i \cdot e_2^i, i\in\{1, \cdots d
\}$) and the other based on clustering the absolute difference of $e_1^i$ and $e_2^i$ ($|e_1^i - e_2^i|, i\in\{1, \cdots d\}$). For both methods, we report the 0th, 25th, 50th, 75th, and 100th percentiles of the energy scores. As shown in Table~\ref{tab:simulation-p1}, our method consistently generates higher energy scores when discriminating content $c_2$, confirming the results outlined in Proposition 1.}

\begin{table}
\centering
\caption{
\textcolor{black}{The energy scores of the masked image features for two different masking approaches. We report the 0th, 25th, 50th, 75th, and 100th percentiles of the energy scores for each approach.}}
\vspace{0.1cm}
\begin{tabular}{lllllll} 
\hline
Masking Proportion & Method                 & 0th      & 25th    & 50th    & 75th    & 100th    \\ 
\hline
5\%                & clustering on $e_1^i \cdot e_2^i$ (Ours) & -10.78 & -6.59 & -5.36 & -4.08 & 1.64   \\
5\%                & clustering on $|e_1^i - e_2^i|$    & -13.87 & -9.63 & -8.56 & -7.54 & -1.60  \\ 
\hline
10\%               & clustering on $e_1^i \cdot e_2^i$ (Ours) & -9.15  & -5.15 & -4.00 & -2.77 & 2.02   \\
10\%               & clustering on $|e_1^i - e_2^i|$    & -11.57 & -8.80 & -7.88 & -7.05 & -2.18  \\ 
\hline
20\%               & clustering on $e_1^i \cdot e_2^i$ (Ours) & -7.46  & -3.46 & -2.36 & -1.35 & 2.66   \\
20\%               & clustering on $|e_1^i - e_2^i|$    & -10.73 & -7.57 & -6.91 & -6.20 & -3.16  \\ 
\hline
30\%               & clustering on $e_1^i \cdot e_2^i$ (Ours) & -6.05  & -2.62 & -1.58 & -0.63 & 2.87   \\
30\%               & clustering on $|e_1^i - e_2^i|$    & -9.19  & -6.73 & -6.13 & -5.59 & -3.31  \\ 
\hline
40\%               & clustering on $e_1^i \cdot e_2^i$ (Ours) & -5.71  & -2.23 & -1.29 & -0.43 & 2.70   \\
40\%               & clustering on $|e_1^i - e_2^i|$    & -8.04  & -5.99 & -5.51 & -5.07 & -3.61  \\ 
\hline
50\%               & clustering on $e_1^i \cdot e_2^i$ (Ours) & -5.24  & -1.94 & -1.13 & -0.37 & 2.69   \\
50\%               & clustering on $|e_1^i - e_2^i|$    & -7.26  & -5.37 & -4.99 & -4.60 & -3.63  \\ 
\hline
60\%               & clustering on $e_1^i \cdot e_2^i$ (Ours) & -4.93  & -1.72 & -0.92 & -0.22 & 2.72   \\
60\%               & clustering on $|e_1^i - e_2^i|$    & -6.23  & -4.85 & -4.53 & -4.19 & -3.29  \\ 
\hline
70\%               & clustering on $e_1^i \cdot e_2^i$ (Ours) & -3.91  & -1.25 & -0.53 & 0.15  & 2.86   \\
70\%               & clustering on $|e_1^i - e_2^i|$    & -5.62  & -4.32 & -4.06 & -3.77 & -2.98  \\ 
\hline
80\%               & clustering on $e_1^i \cdot e_2^i$ (Ours) & -3.11  & -0.77 & -0.14 & 0.53  & 2.20   \\
80\%               & clustering on $|e_1^i - e_2^i|$    & -4.72  & -3.77 & -3.57 & -3.36 & -2.57  \\ 
\hline
90\%               & clustering on $e_1^i \cdot e_2^i$ (Ours) & -2.40  & -0.67 & -0.18 & 0.37  & 2.12   \\
90\%               & clustering on $|e_1^i - e_2^i|$    & -4.00  & -3.32 & -3.15 & -3.01 & -2.55  \\
\hline
\end{tabular}
\label{tab:simulation-p1}
\vspace{-0.4cm}
\end{table}

\begin{figure}
  \centering
  \includegraphics[width=\linewidth]{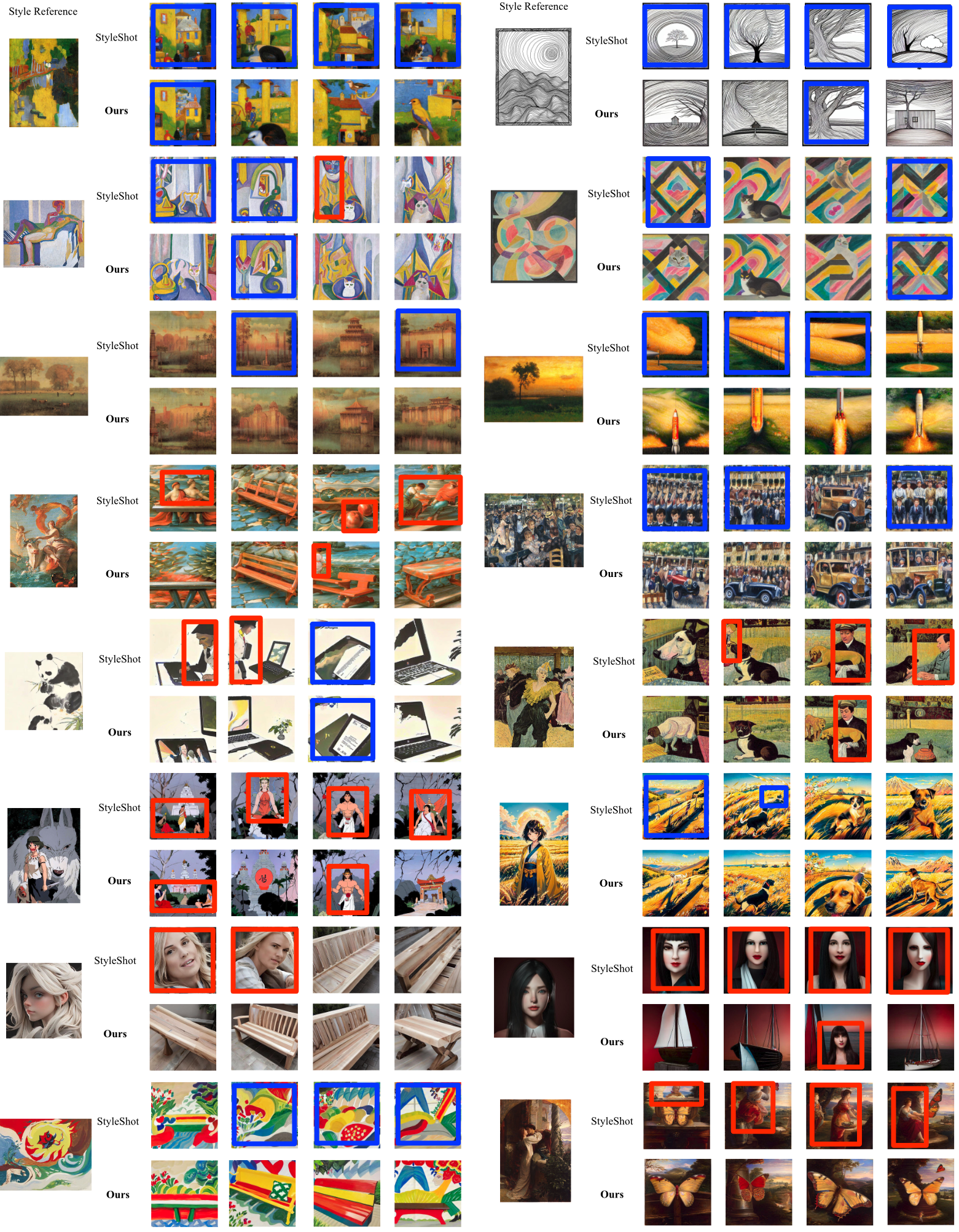}
  \vspace{-0.5cm}
   \caption{
   \textcolor{black}{The one-to-one image generation results between the proposed masking-based method and StyleShot.  We provide multi-sample generalization results for each combination of style reference and target text prompt. 
   To mitigate the influence of randomness, we ensure that all model configurations remain consistent, including the random seed, guidance seed, denoising steps, and other parameters. From the one-to-one comparison, we observe that our method significantly reduces content leakage and alleviates loss of text fidelity, consistently refining StyleShot’s results across all combinations.}
   We mark the results with significant \textcolor{red}{content leakages} and \textcolor{blue}{loss of text fidelity} with \textcolor{red}{red} and \textcolor{blue}{blue} boxes, respectively. 
   }
\label{fig:20}
\end{figure}

\subsection{Visual Comparison in Traditional Style Transfer}

\begin{figure}
  \centering
  \includegraphics[width=0.96\linewidth]{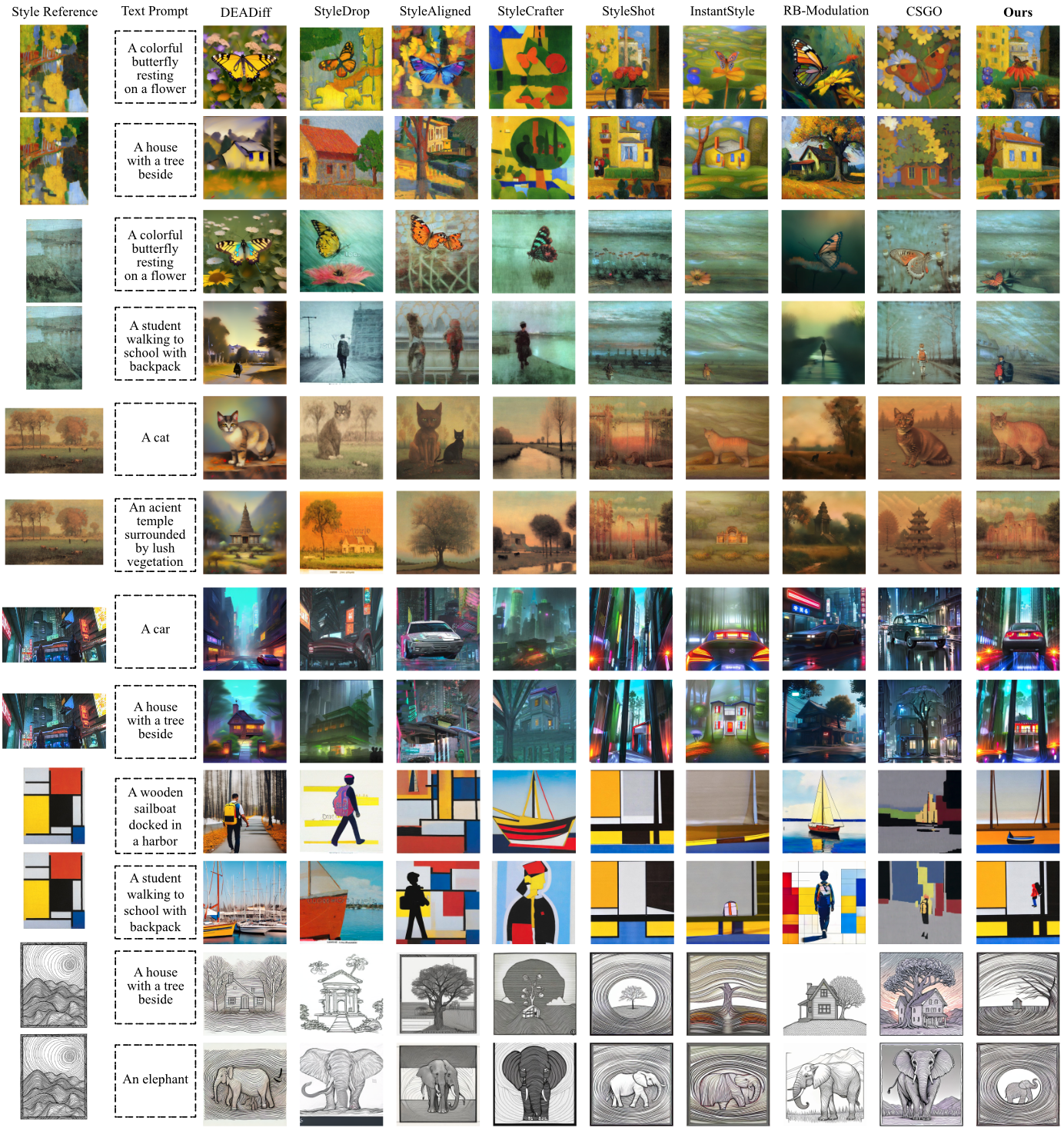}
  \vspace{-0.3cm}
   \caption{
   \textcolor{black}{Visual comparison between recent state-of-the-art methods in traditional style transfer. 
   % The proposed method demonstrates effective stylization, performing better in both object-centered style references and non-object style references.
   }}
   \vspace{-0.2cm}
\label{fig:21}
\end{figure}

\textcolor{black}{In Figure~\ref{fig:21}, we provide additional results comparing the performance of our method with previous approaches in traditional style transfer, where non-object style images are used as style references. As shown in Figure~\ref{fig:21}, when using non-object style references, previous methods such as StyleDrop \citep{sohn2024styledrop} and StyleShot \citep{gao2024styleshot} may suffer from style degradation or a loss of text fidelity. As pointed out in the original paper \citep{rout2024rb}, ``The inherent limitations of the style descriptor or diffusion model might propagate into our framework'', RB-Modulation \cite{rout2024rb} may fail to preserve the style of the reference when the style description does not align well with the image reference, as illustrated in the 3rd and 4th lines in Figure~\ref{fig:21}. As shown in the 9th and 10th lines in Figure~\ref{fig:21}, CSGO \citep{xing2024csgo} may also suffer from style degradation or loss of text fidelity, showing inferior performance compared to our method.
In contrast, from Figure~\ref{fig:5} and Figure~\ref{fig:21}, the proposed method demonstrates superior stylization, performing better in both object-centered style references and non-object style references.
}

\subsection{Ethics Statement}
\label{app-sec:ethics}

\textcolor{black}{This work aims to make a positive impact on the field of AI-driven image generation. We aim to facilitate the creation of images with diverse styles, but we expect all related processes to comply with local laws and be used responsibly.}

\begin{figure}
  \centering
  \includegraphics[width=\linewidth]{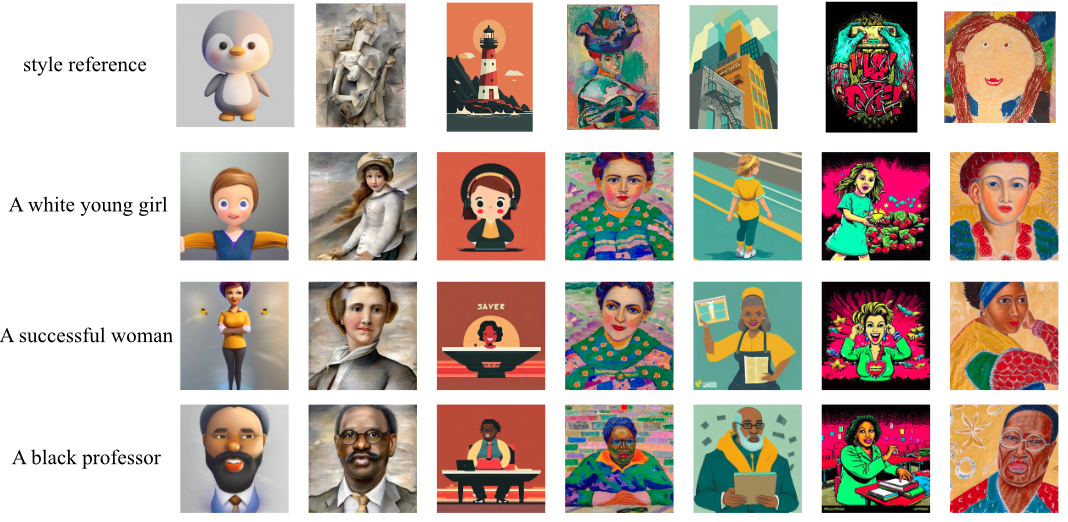}
  \vspace{-0.5cm}
   \caption{\textcolor{black}{Examples of text prompts contain different genders, skin colors, and ages.}}
   \vspace{-0.5cm}
\label{fig:22}
\end{figure}

\textcolor{black}{
The use of AI to generate human-related images, particularly those involving characteristics such as skin color, gender, age, and other demographic factors, raises complex ethical questions. We are aware that the generation of images involving these attributes must be handled with care to avoid reinforcing stereotypes, perpetuating discriminations, or contributing to the misrepresentations of certain groups. We take these concerns very seriously and believe that AI should be used in a way that promotes fairness, inclusion, and respect for all individuals. 
Here, we give several examples of text prompts containing different genders, skin colors, and ages, as shown in Figure~\ref{fig:22}.
% In light of these considerations, the code and methodology in this paper shall be used responsibly. The use of this material of this paper shall avoid any potential bias related to sensitive attributes such as gender, race, and age, etc.
}

\textcolor{black}{We observe that in most cases, our method is able to generate images with diversity. However, there are certain cases that general image generation methods can be misused. In light of these considerations, we emphasize that the code and methodology presented in this paper must be used responsibly. Users are expected to utilize this material in a way that avoids any potential bias related to sensitive attributes such as gender, race, age, and other demographic factors. We believe that the responsible use of AI-driven image generation tools is essential to fostering ethical and equitable outcomes in the field.}

\end{document}